%% file: workshop_paper.tex
\documentclass[10pt,twocolumn,letterpaper]{article}

\usepackage{cvpr}      %

\input{preamble}
\usepackage{tabularx}  %

\definecolor{cvprblue}{rgb}{0.21,0.49,0.74}
\usepackage[pagebackref,breaklinks,colorlinks,allcolors=cvprblue]{hyperref}

\def\paperID{11} %
\def\confName{CVPR}
\def\confYear{2026}

\title{Geometry-Guided Camera Motion Understanding in VideoLLMs}

\author{Haoan Feng\textsuperscript{1},
Sri Harsha Musunuri\textsuperscript{2},
Guan-Ming Su\textsuperscript{2},
\\
\textsuperscript{1}University of Maryland, College Park,\quad
\textsuperscript{2}Dolby Laboratories Inc. \\\\
{\tt\small hfengac@umd.edu},\quad
{\tt\small harsha.musu@gmail.com},\quad
{\tt\small guanmingsu@ieee.org}\quad
}

\begin{document}
\maketitle

\input{sec/0_abstract}
\input{sec/1_Introduction}
\input{sec/2_Relatedwork}
\input{sec/3_Method}
\input{sec/4_Experiment}

\input{sec/5_Conclusion}

\section*{Acknowledgments}
This work was conducted during the first author's internship at Dolby Laboratories Inc. Special thanks to Xinran Wang~\cite{wang2025cinetechbench} for sharing the full CineTechBench dataset.

{
    \small
    \bibliographystyle{ieeenat_fullname}
    \bibliography{main}
}

\end{document}


\maketitle

\appendix
\tableofcontents

\section{Camera Motion Taxonomy and Constraints}
\label{sec:supp_taxonomy}

We adopt the camera motion taxonomy proposed in CameraBench~\cite{lin2025towards} and refer readers to that work for a comprehensive definition of motion primitives, annotation guidelines, and constraint design. CameraBench formalizes camera motion as a set of atomic cinematographic operations (\eg, pan, tilt, dolly, roll, arc) with clear geometric interpretation and explicit rules to avoid contradictory labels.

In this work, we use a controlled subset of 15 primitives tailored to short (1-second) within-shot segments, focusing on the extrinsic parameter changes of the camera. Specifically, we focus on principal rotation and translation operations that can be reliably identified at this temporal granularity, while preserving the incompatibility structure defined in CameraBench (\eg, opposing directions along the same degree of freedom are mutually exclusive, and \texttt{\small static} cannot co-occur with any non-static motion).

Our taxonomy is therefore not a redefinition, but a task-specific instantiation of the CameraBench framework, designed to support constrained multi-label benchmarking and systematic diagnosis of camera motion recognition in VideoLLMs.

\begin{figure}[t]
\centering
\includegraphics[width=\columnwidth]{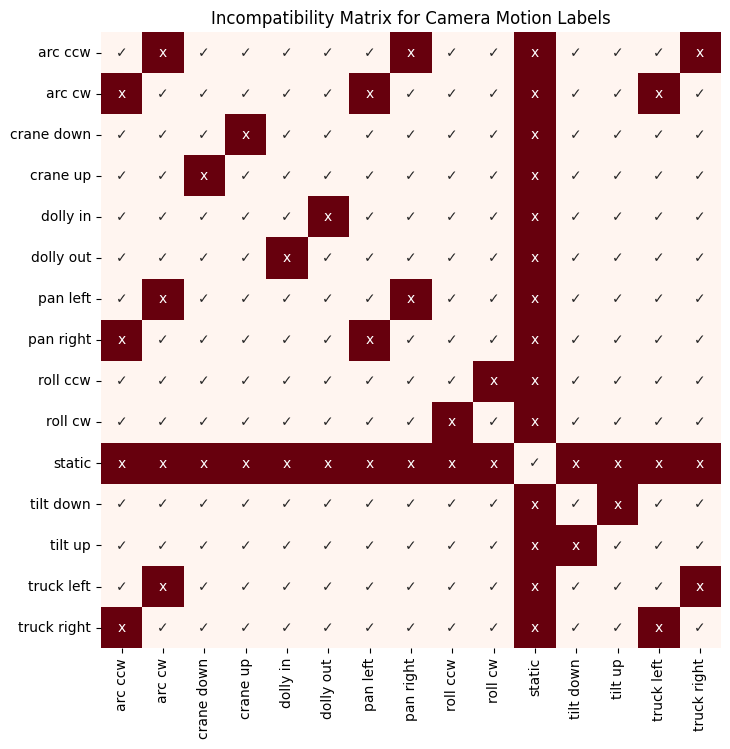}
\caption{\textbf{Incompatibility matrix for constrained multi-label camera motion.}
Entry $\mathbf{M}_{ij}{=}1$ indicates that primitives $i$ and $j$ are mutually exclusive and cannot co-occur in the same 1-second segment.
The axis structure produces block-wise exclusivity (\eg, left/right pairs), and \texttt{\small static} is incompatible with all non-static primitives.
Row/column ordering matches the primitive index order used in our dataset and experiments.}
\label{fig:supp_incompat}
\end{figure}

\subsection{Atomic Motion Primitives}
\label{sec:supp_primitives}
Our benchmark instantiates a compact, implementation-friendly subset of $K{=}15$ atomic motion primitives, summarized in \cref{tab:supp_primitives}. 
Each primitive corresponds to a distinct geometric degree of freedom of the camera motion: three rotations (yaw, pitch, roll), three translations (X, Y, Z), a coupled rotation--translation pattern (\texttt{\small arc}), and a \texttt{\small static} state.
This axis-aligned parameterization keeps the label space interpretable, physically grounded, and directly compatible with 3D camera representations.

The primitives are designed to be identifiable at the 1-second, within-shot granularity used in our benchmark, and they follow standard cinematographic terminology.
Across different axes, primitives may co-occur (\eg, \texttt{\small pan left} + \texttt{\small tilt up}, or \texttt{\small dolly in} + \texttt{\small pan right}), motivating a multi-label rather than single-label formulation.
We also include \texttt{\small arc} to capture common cinematographic motion where yaw rotation is coupled with lateral translation, which is difficult to describe as a single pure-axis primitive.
Finally, \texttt{\small static} denotes the absence of dominant camera motion above predefined thresholds and serves as a diagnostic case for distinguishing true camera motion from object-induced or scene-induced apparent motion in VideoLLMs.

\begin{table*}[t]
\centering
\small
\setlength{\tabcolsep}{4pt}
\caption{\textbf{Atomic camera motion primitives.}
We list the $K{=}15$ primitives used in our benchmark, grouped by geometric degree of freedom (rotations, translations, coupled motion, and static).
Opposing directions within each axis are mutually exclusive, forming the basis of our incompatibility constraints, while primitives across different axes may co-occur under a constrained multi-label formulation.}\label{tab:supp_primitives}
\begin{tabular}{llll}
\toprule
\textbf{Axis} & \textbf{Primitive} & \textbf{Operational Definition} & \textbf{Mutual exclusion} \\
\midrule
Yaw & pan left / pan right & Horizontal rotation about vertical axis & pan left $\leftrightarrow$ pan right \\
Pitch & tilt up / tilt down & Vertical rotation about horizontal axis & tilt up $\leftrightarrow$ tilt down \\
Roll & roll cw / roll ccw & In-plane rotation about optical axis & roll cw $\leftrightarrow$ roll ccw \\
Y & truck left / truck right & Lateral translation parallel to image plane & truck left $\leftrightarrow$ truck right \\
Z & crane up / crane down & Vertical translation of camera center & crane up $\leftrightarrow$ crane down \\
X & dolly in / dolly out & Translation along view axis & dolly in $\leftrightarrow$ dolly out \\
Coupled & arc cw / arc ccw & Coupled yaw + lateral translation & arc cw $\leftrightarrow$ arc ccw \\
None & static & No dominant motion above thresholds & static $\leftrightarrow$ any non-static \\
\bottomrule
\end{tabular}
\end{table*}

\subsection{Incompatibility constraints and label-set canonicalization}
\label{sec:supp_constraints}
Camera motion primitives are not independent.
We enforce geometric exclusivity constraints using a symmetric incompatibility matrix
$\mathbf{M} \in \{0,1\}^{K \times K}$, where $\mathbf{M}_{ij}{=}1$ indicates that primitives $i$ and $j$ cannot co-occur in the same 1-second segment.
\cref{fig:supp_incompat} visualizes $\mathbf{M}$; the row/column ordering matches the primitive index ordering used throughout the dataset, training, and evaluation. Within each degree of freedom, opposing directions are mutually exclusive (\eg, \texttt{\small pan left} vs.\ \texttt{\small pan right}), and \texttt{\small static} is mutually exclusive with any non-static primitive.
These constraints prevent contradictory supervision and yield a well-defined target space for constrained multi-label recognition (and the corresponding VQA-style evaluation).

To make labels deterministic and comparable across data generation, training logs, and evaluation scripts, we canonicalize each segment's label set as follows:
(i) remove duplicates; 
(ii) drop any label set that violates $\mathbf{M}$ (\ie, contains a conflicting pair);
(iii) sort remaining labels by a fixed global order (the primitive index order used in \cref{fig:supp_incompat});
and (iv) enforce the cardinality of at least one and at most three primitives per segment.
The last step reflects our benchmark design assumption that 1-second clips rarely contain more than three dominant, distinguishable camera motions, and it ensures a bounded and interpretable output space for both classifiers and VideoLLM prompting.

\section{Dataset Construction Details}

We construct \textbf{CameraMotionDataset} and \textbf{CameraMotionVQA} to provide controlled, geometry-aware supervision and standardized evaluation for camera motion understanding in VideoLLMs.
As summarized in \cref{tab:positioning_full}, existing benchmarks primarily focus on broad cinematographic techniques, perceptual composition attributes, or real-world camera motion without explicit geometric supervision.
In contrast, our dataset is built from synthetic videos with known camera parameters, segmented into fixed 1-second within-shot clips, and annotated with constrained multi-label motion primitives.
As shown in the main paper, this design enables (i) precise primitive-level supervision, (ii) constraint-consistent evaluation via a multiple-choice VQA protocol, and (iii) controlled studies of geometry-derived camera cues and motion-token distillation.
Below, we detail the data source, segmentation strategy, label generation process, and balancing procedure.

\begin{table*}[t]
\centering
\scriptsize
\setlength{\tabcolsep}{2pt}
\renewcommand{\arraystretch}{1.15}
\caption{Positioning of CameraMotionDataset and CameraMotionVQA against related benchmarks and cinematic datasets. We contrast task focus, temporal unit, label structure (including whether explicit constraints or QA protocols are provided), camera-motion granularity, intended use, and whether explicit camera parameters (geometry access) are available.}\label{tab:positioning_full}
\begin{tabular}{p{2.3cm}|p{2.3cm}|p{2.3cm}|p{2.3cm}|p{2.3cm}|p{2.3cm}|p{2.1cm}}
\hline
\textbf{Work} & \textbf{Primary focus} & \textbf{Temporal unit} & \textbf{Label type (constraints / QA)} & \textbf{Motion granularity} & \textbf{Intended use} & \textbf{Geometry access} \\
\hline
CameraBench~\cite{lin2025towards} & Camera motion primitives in diverse real-world videos; taxonomy + label-then-caption annotations. & Short video clips (avg. \mbox{\textasciitilde}5--6s); shot-aligned / manually segmented. & Multi-label binary primitive tags (\mbox{\textasciitilde}50); plus captions and paired yes/no VQA skill probes. & Primitive-level; directional primitives (translation / rotation / zoom / tracking, etc.). & Mixed: benchmark suite (classification, retrieval, captioning, VQA) + small-scale training for motion-aware VLMs. & \textbf{No}: real videos (no ground-truth camera parameters). \\
\hline
CineTechBench~\cite{wang2025cinetechbench} & Broad cinematographic techniques (scale, angle, composition, movement, lighting, color, focal length) with expert annotation. & Movie images + short movie clips/shots (\mbox{\textasciitilde}600 images, 120 clips). & Manual technique labels; question--answer pairs and annotated descriptions for understanding; separate camera-movement generation evaluation. & Coarse-to-mid: technique categories; camera movement is one dimension among 7. & Diagnostic benchmark (understanding + generation evaluation). & \textbf{No}: no explicit camera parameters provided. \\
\hline
VidComposition~\cite{tang2025vidcomposition} & Composition understanding in compiled videos (cinematography, character, narrative, scene, making). & Compiled videos (avg. \mbox{\textasciitilde}20 min) segmented into coherent sections. & Human-annotated multiple-choice QA (982 videos / 1706 questions) across 15 sub-tasks, including camera movement / angle / shot size perception. & Coarse-to-mid: composition attributes; camera motion appears as a perceptual attribute in QA. & Diagnostic benchmark for MLLMs on compiled-video composition. & \textbf{No}: no explicit camera parameters provided. \\
\hline
CineScale2~\cite{savardi2023cinescale2} & Cinematic camera features (camera angle + camera level) on movie frames/images. & Frames / images (\mbox{\textasciitilde}24.7k). & Single-label categorical annotations per attribute (angle: 5 classes; level: 6 classes). & \emph{N/A for motion} (static camera setup attributes only). & Training-scale dataset for recognition of camera angle/level; film-style analysis. & \textbf{No}: no explicit camera parameters provided. \\
\hline
\textbf{CameraMotionDataset} & Camera motion primitive dataset for controlled supervision and motion-token distillation. & \textbf{Within-shot 1s segments} (fixed-length; shot-consistent). & \textbf{Constrained multi-label} primitive annotations with axis-wise incompatibility constraints. & \textbf{Primitive-level; directional; axis-wise constraints}; supports compound motions within 1s. & \textbf{Supervision dataset} (training / distillation / controlled analysis). & \textbf{Yes}: synthetic camera parameters available (controlled rendering). \\
\hline
\textbf{CameraMotionVQA} & Camera-motion sensitivity benchmark for VideoLLM evaluation via structured prompting. & Same 1s within-shot segments mapped to QA format. & \textbf{Standardized multiple-choice VQA protocol} derived from ground-truth primitives; constraint-consistent answer space. & Primitive-level reasoning over directional motions (single and compound). & \textbf{Diagnostic benchmark} for motion-aware understanding and prompting. & Implicit (inherits synthetic geometry from CameraMotionDataset; not directly exposed to models). \\
\hline
\end{tabular}
\end{table*}

\subsection{Pose-to-label mapping}
\label{sec:supp_pose2label}

Each 1-second segment contains $T{=}15$ camera extrinsics
$\{ \mathbf{E}_t \}_{t=1}^{T}$,
where $\mathbf{E}_t = [\mathbf{R}_t \mid \mathbf{t}_t] \in \mathbb{R}^{3 \times 4}$.
Translations are expressed in world units (meters in the renderer), and rotations are represented as $3{\times}3$ matrices.
We assume a right-handed coordinate system where the camera forward axis corresponds to the negative $x$-axis, and yaw rotation occurs about the world vertical axis.

The pose sequence is mapped to primitive labels using two signals: (i) net translation expressed in the initial camera frame and (ii) accumulated inter-frame rotation. All angles are computed in degrees, and Algorithm~\ref{alg:pose2label} summarizes the full procedure.

\paragraph{Rotation statistics.}
For consecutive frames, we compute the relative rotation
$\mathbf{R}_{\Delta,t}=\mathbf{R}_{t}\mathbf{R}_{t-1}^\top$
and convert it to an axis--angle representation.
We accumulate signed changes for yaw (pan), pitch (tilt, estimated from the forward-vector change), and roll.
Let $\Delta_{\text{pan}}, \Delta_{\text{tilt}}, \Delta_{\text{roll}}$ denote the total signed changes over the segment, and $\Sigma_{\text{pan}}, \Sigma_{\text{tilt}}, \Sigma_{\text{roll}}$ denote the accumulated absolute changes.

\noindent\textbf{Static / rotation-dominant case.}
Let $d_{\text{trans}}$ be the total translation distance (sum of inter-frame translations).
If $d_{\text{trans}} < 0.05$, we treat the segment as rotation-dominant.
We then assign \texttt{\small static} if $\Sigma_{\text{pan}}<0.2$ and $\Sigma_{\text{tilt}}<0.2$; otherwise we assign \texttt{\small pan left/right} based on the sign of $\Delta_{\text{pan}}$ when $\Sigma_{\text{pan}}>0.2$, and \texttt{\small tilt up/down} based on the sign of $\Delta_{\text{tilt}}$ when $\Sigma_{\text{tilt}}>0.2$.

\noindent\textbf{Translation-dominant case.}
If translation is significant, we compute the net translation in the initial camera frame:
\[
\Delta \mathbf{t}_{\text{cam}}=\mathbf{R}_1^\top(\mathbf{t}_T-\mathbf{t}_1),
\]
with components $(x,y,z)$ corresponding to forward, lateral, and vertical motion, respectively.
We use an adaptive threshold
$t_{\text{move}}=\max\!\big(0.3\cdot\|\Delta \mathbf{t}_{\text{cam}}\|_\infty,\;0.5\big)$
to decide whether a component is dominant.

To distinguish \texttt{\small arc} from near-straight translation, we estimate path curvature
\[
\kappa=\frac{\sum_t \|\mathbf{f}_{t}-\mathbf{f}_{t-1}\|_2}{\sum_t \|\mathbf{t}_{t}-\mathbf{t}_{t-1}\|_2+\epsilon},
\]
where $\mathbf{f}_t$ is the camera forward vector in world coordinates.
If $\kappa>9{\times}10^{-4}$, we classify the segment as \texttt{\small arc cw/ccw} based on the sign of $\Delta_{\text{pan}}$, and optionally add auxiliary \texttt{\small tilt} or \texttt{\small dolly} if their corresponding rotation/translation magnitudes exceed thresholds.
Otherwise, we assign axis-aligned translation primitives according to dominant components: \texttt{\small dolly in/out} from $x$, \texttt{\small truck left/right} from $y$, and \texttt{\small crane up/down} from $z$.

\noindent\textbf{Roll.}
We additionally assign \texttt{\small roll cw/ccw} only when forward motion is present and $\Sigma_{\text{roll}} > 0.5$, which avoids spurious roll labels from minor numerical jitter.

To verify the correctness of the automatic mapping, we randomly sampled 720 segments and collected primitive annotations using the same 15-label set.
Annotators were instructed to select the correct option from an MCQ-style VQA benchmark and to prioritize dominant motion over minor jitter.
The resulting agreement accuracy is 93\%, supporting the reliability of the pose-to-label mapping procedure.

\begin{algorithm}[t]
\caption{Pose-to-label mapping}
\label{alg:pose2label}
\footnotesize
\begin{algorithmic}[1]
\Require Extrinsics $\{(\mathbf{R}_t,\mathbf{t}_t)\}_{t=1}^{T}$
\State $\mathcal{Y}\gets \emptyset$
\State Compute $d_{\text{trans}}$ (total translation distance) and $(\Delta_{\text{pan}},\Delta_{\text{tilt}})$, $(\Sigma_{\text{pan}},\Sigma_{\text{tilt}})$ (signed / abs rotation stats)
\If{$d_{\text{trans}} < 0.05$} \Comment{rotation-dominant / static}
    \If{$\Sigma_{\text{pan}} < 0.2$ \textbf{and} $\Sigma_{\text{tilt}} < 0.2$}
        \State \Return \{\texttt{static}\}
    \EndIf
    \If{$\Sigma_{\text{pan}} > 0.2$}
        \State $\mathcal{Y}\gets \mathcal{Y}\cup\{\texttt{pan right}\}$ if $\Delta_{\text{pan}}>0$ else $\mathcal{Y}\gets \mathcal{Y}\cup\{\texttt{pan left}\}$
    \EndIf
    \If{$\Sigma_{\text{tilt}} > 0.2$}
        \State $\mathcal{Y}\gets \mathcal{Y}\cup\{\texttt{tilt down}\}$ if $\Delta_{\text{tilt}}>0$ else $\mathcal{Y}\gets \mathcal{Y}\cup\{\texttt{tilt up}\}$
    \EndIf
\Else \Comment{translation-dominant}
    \State Compute $\Delta\mathbf{t}_{\text{cam}}$ and $t_{\text{move}}$
    \State Compute curvature $\kappa$ from forward vectors
    \If{$\kappa > 9\times 10^{-4}$}
        \State $\mathcal{Y}\gets \mathcal{Y}\cup\{\texttt{arc cw}\}$ if $\Delta_{\text{pan}}>0$ else $\mathcal{Y}\gets \mathcal{Y}\cup\{\texttt{arc ccw}\}$
    \EndIf
    \State Update $\mathcal{Y}$ with \texttt{dolly in/out} (forward-dominant), \texttt{truck left/right} (lateral-dominant), and \texttt{crane up/down} (vertical-dominant)
\EndIf
\State \Return Canonicalize($\mathcal{Y}$) (Sec.~\ref{sec:supp_constraints})
\end{algorithmic}
\end{algorithm}

\subsection{Re-balancing and class distribution}
\label{sec:supp_rebalance}

\begin{figure*}[t]
\centering
\includegraphics[width=0.9\textwidth]{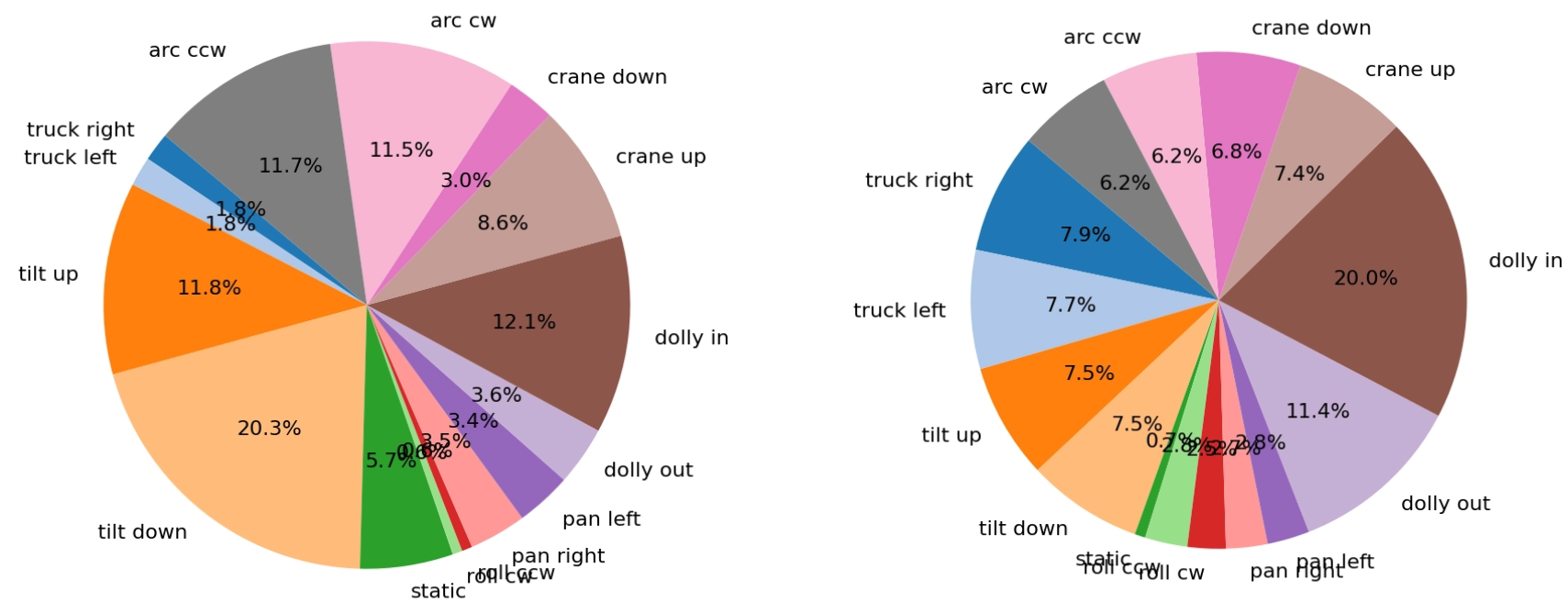}
\caption{\textbf{Label-set distribution before and after re-balancing.}
We cap each canonical primitive combination at 200 segments.
The original distribution is highly skewed toward common directional motions, while the rebalanced dataset provides a more uniform coverage across primitive combinations.}
\label{fig:supp_distribution}
\end{figure*}

The full pose-to-label conversion over all rendered videos produces 542{,}504 1-second segments.
As shown in \cref{fig:supp_distribution}(a), due to the scripted camera trajectories in the synthetic generator, the resulting label distribution is highly imbalanced.
In particular, simple directional primitives such as \texttt{\small tilt down} and \texttt{\small arc ccw} occur far more frequently than rarer motions such as \texttt{\small roll cw/ccw}.
Without re-balancing, this imbalance would bias classifiers toward dominant primitives and obscure diagnostic conclusions about motion sensitivity in VideoLLMs.

To mitigate this effect, we apply per-label-set downsampling.
For each unique canonical label set (Sec.~\ref{sec:supp_constraints}) (\eg, \texttt{\small tilt up} and \texttt{\small pan left}), we cap the number of segments to at most 200 samples.
If a label set contains fewer than 200 segments, all instances are retained.
This strategy preserves the diversity of primitive combinations while preventing over-representation of frequent motion patterns.

After re-balancing, the final dataset contains \textbf{12{,}274} segments.
\cref{fig:supp_distribution} visualizes the label-set frequency distribution before and after downsampling.
The post-balanced distribution is substantially flatter, enabling more controlled evaluation of primitive-level recognition and structured prompting.

\section{CameraMotionVQA Protocol}
\label{sec:supp_vqa}

\subsection{Prompt templates and answer normalization}
\label{sec:supp_vqa_prompt}

Each 1-second clip in CameraMotionDataset is converted into a 4-way multiple-choice VQA instance by pairing the video with a fixed prompt template and four candidate label sets (one correct and three distractors). We use the following prompt string (shown verbatim) with lettered options:
\begin{quote}\footnotesize
\texttt{<video>\\
Identify the camera motion depicted in the video using standard cinematographic terminology.\\
Options:\\
(A) <optA>\\
(B) <optB>\\
(C) <optC>\\
(D) <optD>}
\end{quote}
The reference answer is stored as \texttt{\footnotesize Answer: <letter>}.

\paragraph{Option text.}
For readability, we verbalize \texttt{cw/ccw} as \texttt{\small clockwise/counterclockwise} in options, while preserving all other primitive names (\eg, \texttt{\small pan left}, \texttt{\small dolly in}). Multi-primitive options are formatted as a comma-separated list in canonical order (Sec.~\ref{sec:supp_constraints}).

\paragraph{Answer parsing.}
At evaluation time, we normalize model outputs by extracting the first occurrence of a valid choice letter in \{\texttt{A},\texttt{B},\texttt{C},\texttt{D}\} (case-insensitive), optionally preceded by the token \texttt{\small Answer:}. If no valid letter is found, model output will be parsed to try to match the ground truth; otherwise, the prediction is marked as invalid. This normalization ensures a consistent, model-agnostic scoring interface for VideoLLMs.

\subsection{Distractor sampling algorithm}
\label{sec:supp_vqa_distractors}

As shown in Algorithm~\ref{alg:distractors}, a na\"ive distractor strategy (sampling arbitrary label sets) can make questions either trivial (\eg, mismatched label cardinality) or invalid (\eg, violating incompatibility constraints). We therefore sample distractors from a precomputed pool $\mathcal{Y}$ of \emph{canonicalized, constraint-valid} label sets (Sec.~\ref{sec:supp_constraints}), stratified by label cardinality $c\in\{1,2,3\}$.

\paragraph{Complexity-matched distractors.}
Given a ground-truth label set $y^*$ with complexity $c=|y^*|$, we sample distractors using the following rule, matching the implementation in our dataset generator:
\begin{itemize}\setlength\itemsep{0pt}
\item If $c=1$: sample 3 distractors from single-label sets and 1 distractor from two-label sets.
\item If $c=2$: sample 2 distractors from two-label sets, 1 from single-label sets, and 1 from three-label sets.
\item If $c=3$: sample 2 distractors from two-label sets and 1 from three-label sets (then insert $y^*$ to ensure one correct option).
\end{itemize}
This yields a mixture of easy and moderately hard negatives while avoiding degenerate cases where the correct answer is identifiable solely by the number of primitives.

\paragraph{Constraint validity.}
All candidates (including distractors) are drawn from $\mathcal{Y}$, which contains only incompatibility-consistent label sets. As a result, every option is valid under the same constraint system as the ground truth, preventing models from exploiting constraint violations as shortcuts.

\begin{algorithm}[t]
\caption{Distractor sampling for CameraMotionVQA (4-way MCQ)}
\label{alg:distractors}
\footnotesize
\begin{algorithmic}[1]
\Require GT label set $y^*$; pools $\mathcal{Y}_1,\mathcal{Y}_2,\mathcal{Y}_3$ of valid label sets with $|y|=1,2,3$
\State $c\gets |y^*|$; $\mathcal{O}\gets \emptyset$
\If{$c=1$}
    \State $\mathcal{O}\gets$ Sample$(\mathcal{Y}_2,1)\ \cup\ $Sample$(\mathcal{Y}_1,3)$
\ElsIf{$c=2$}
    \State $\mathcal{O}\gets$ Sample$(\mathcal{Y}_1,1)\ \cup\ $Sample$(\mathcal{Y}_3,1)\ \cup\ $Sample$(\mathcal{Y}_2,2)$
\Else
    \State $\mathcal{O}\gets$ Sample$(\mathcal{Y}_2,2)\ \cup\ $Sample$(\mathcal{Y}_3,1)$
\EndIf
\State Replace one element in $\mathcal{O}$ with $y^*$ to ensure exactly one correct option
\State Verbalize \texttt{cw/ccw} $\rightarrow$ \texttt{clockwise/counterclockwise}; shuffle and assign letters A--D
\State \Return Options $\{(A,y_A),(B,y_B),(C,y_C),(D,y_D)\}$ and correct letter
\end{algorithmic}
\end{algorithm}

\section{Additional Experimental Results}
\subsection{Classifier capacity ablation}
\label{sec:supp_classifier}

We study the effect of classifier capacity on camera-motion recognition.
The classifier is a lightweight Transformer encoder operating on frozen VideoLLM vision features.
We vary three architectural parameters: the number of encoder blocks (2--6), the number of attention heads (4 or 8), and the hidden dimension (256--896).

\begin{figure}[t]
    \centering
    \includegraphics[width=\linewidth]{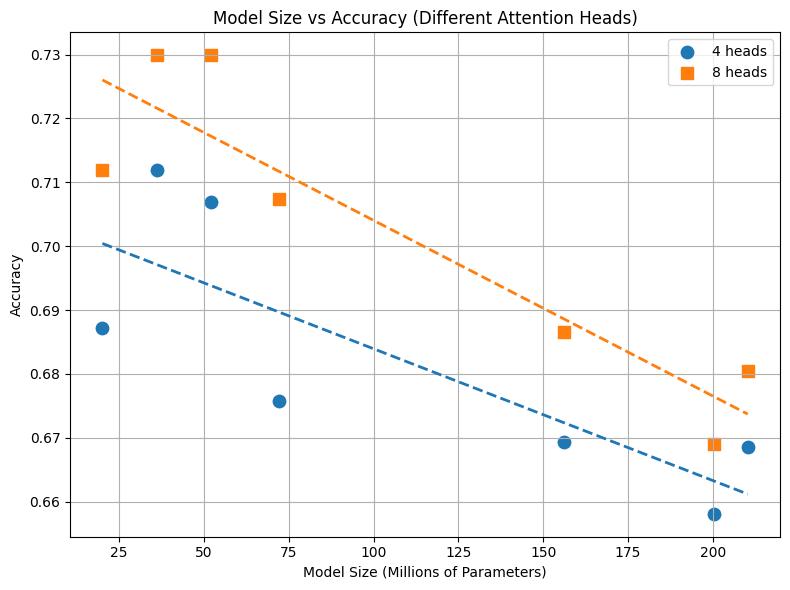}
    \caption{\textbf{Classifier capacity ablation.} Accuracy versus classifier size (M parameters) for varying Transformer depth, head count, and hidden dimension. Overall performance saturates with small models; using 8 attention heads yields a modest but consistent gain over 4 heads.}
    \label{fig:supp_modelsize}
\end{figure}

\cref{fig:supp_modelsize} plots accuracy as a function of model size.
Across all configurations, performance varies within a relatively narrow range (about 0.71--0.79), indicating that increased classifier capacity does not consistently translate to higher accuracy.
The best results are obtained with small models (20--50M parameters), while larger configurations with wider hidden dimensions or additional layers do not provide reliable improvements.

Increasing the number of attention heads yields a modest gain (typically 1--2\%), but the overall trend shows early saturation: models larger than roughly 50M parameters do not improve performance and sometimes degrade slightly.
This behavior is consistent with a representation bottleneck in the frozen vision features, but it may also reflect limited supervision after re-balancing (Sec.~\ref{sec:supp_rebalance}), where the effective training set size is relatively small for fitting larger classifiers.
A potential direction to better exploit higher-capacity heads is to increase data diversity via augmentation, \eg, temporal replay/resampling within the 1-second clip, synthetic camera roll perturbations, and scale-preserving random crops (to simulate focal-length and framing variation) while keeping the motion label unchanged.

\subsection{Per-label breakdown and confusion analysis}
\label{sec:supp_confusion}

We provide per-label analysis for the three camera-motion recognition pipelines used throughout the paper: \emph{VGGT classifier} (teacher cues), \emph{VGGT--Q-Former} (distilled student cues), and \emph{Q-Former probing} (diagnostic baseline on frozen VideoLLM vision features).
Following Tab.~2 in the main paper, we report class-averaged and frequency-weighted scores: the overall \textbf{Macro-F1 / Weighted-F1} are
\textbf{0.87 / 0.92} (VGGT classifier),
\textbf{0.83 / 0.87} (VGGT--Q-Former),
and \textbf{0.69 / 0.74} (Q-Former probing).
These aggregate gaps are reflected consistently in the per-primitive results in \cref{tab:supp_perlabel}.

\begin{table}[t]
\centering
\small
\setlength{\tabcolsep}{6pt}
\caption{\textbf{Per-primitive F1 scores.}
VGGT classifier achieves the best accuracy and consistently outperform the other two methods. A similar trend is visible between VGGT--Q-Former and Q-Former probing. However, the VGGT classifier requires running the 3D backbone, leading to higher computational and memory overhead.
VGGT--Q-Former reuses frozen VideoLLM vision features and trades a modest accuracy drop for substantially higher throughput.
Compared to probing, both VGGT-derived variants improve nearly all primitives, with the largest gains on ambiguous motions (\texttt{pan}, \texttt{tilt}) and the \texttt{static} class.}
\label{tab:supp_perlabel}
\begin{tabular}{lccc}
\toprule
\textbf{Primitive} & \shortstack{\textbf{VGGT}\\\textbf{Classifier}} &
\shortstack{\textbf{VGGT--}\\\textbf{Q-Former}} &
\shortstack{\textbf{Q-Former}\\\textbf{Probing}} \\
\midrule
arc ccw & 0.88 & 0.83 & 0.68 \\
arc cw & 0.94 & 0.90 & 0.75 \\
crane down & 0.94 & 0.89 & 0.74 \\
crane up & 0.93 & 0.88 & 0.72 \\
dolly in & 0.94 & 0.90 & 0.78 \\
dolly out & 0.90 & 0.85 & 0.75 \\
pan left & 0.83 & 0.76 & 0.58 \\
pan right & 0.85 & 0.80 & 0.66 \\
roll ccw & 0.91 & 0.86 & 0.71 \\
roll cw & 0.97 & 0.92 & 0.83 \\
static & 0.65 & 0.44 & 0.27 \\
tilt down & 0.94 & 0.90 & 0.78 \\
tilt up & 0.89 & 0.83 & 0.69 \\
truck left & 0.93 & 0.88 & 0.74 \\
truck right & 0.92 & 0.88 & 0.77 \\
\midrule
\textbf{Macro F1} & 0.87 & 0.83 & 0.69 \\
\bottomrule
\end{tabular}
\end{table}

\paragraph{Per-label breakdown.}
\cref{tab:supp_perlabel} reports F1 for each primitive.
VGGT-derived cues yield uniformly strong performance across most translation and rotation primitives, with several classes exceeding 0.9 F1 (\eg, \texttt{\small dolly in/out}, \texttt{\small crane up/down}, \texttt{\small truck left/right}, \texttt{\small roll cw}).
Distilling VGGT cues into a compact Q-Former incurs a moderate but systematic drop (typically 0.03--0.06 F1 per class), while still substantially outperforming probing.
This aligns with the intended accuracy--throughput trade-off: the VGGT classifier achieves the best recognition accuracy but requires running the 3D backbone, whereas VGGT--Q-Former reuses frozen VideoLLM visual features at inference time and is therefore significantly more efficient.

The most challenging class is \texttt{\small static}, which has the lowest F1 across all methods.
This is expected because \texttt{\small static} is defined by the \emph{absence} of dominant motion above thresholds, and minor camera jitter or object-induced apparent motion can easily trigger false positives.
Similarly, \texttt{\small pan} and \texttt{\small tilt} are moderately harder than most translation primitives, as they are more easily confounded with arc-like camera paths and with subtle changes in viewpoint.

\begin{figure*}[t]
    \centering
    \includegraphics[width=0.75\linewidth]{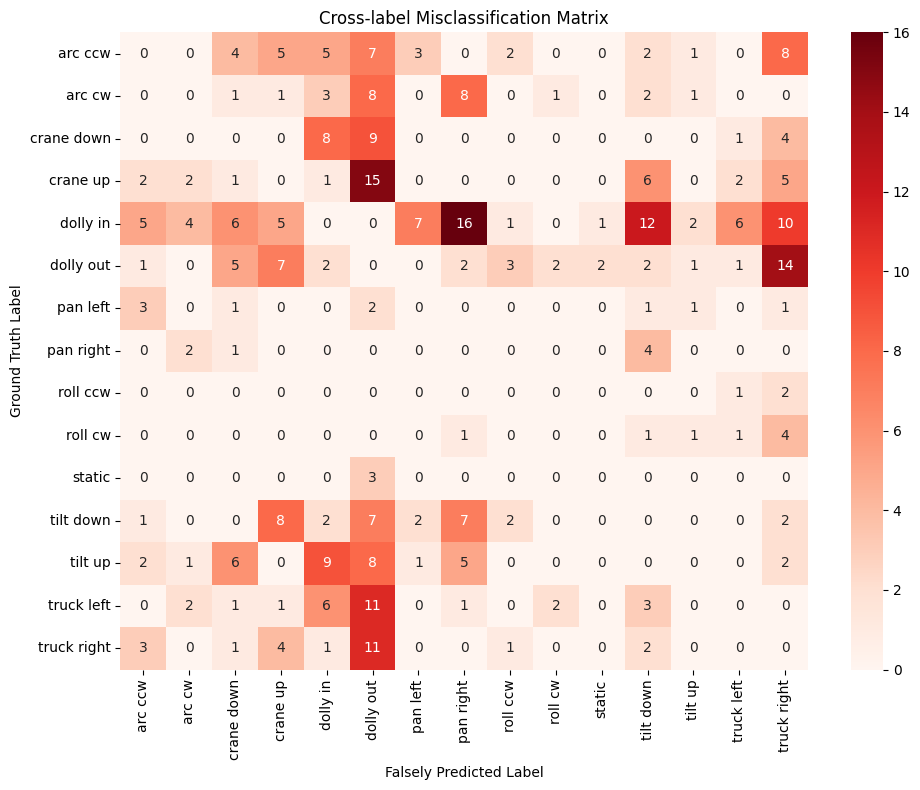}
    \caption{\textbf{Cross-label confusion for the VGGT classifier.}
    Rows denote ground-truth primitives and columns denote falsely predicted primitives (higher intensity indicates more frequent confusion).
    The label order matches \cref{tab:supp_perlabel} and the dataset primitive indices.}
    \label{fig:supp_confusion}
\end{figure*}

\paragraph{Confusion analysis.}
\cref{fig:supp_confusion} reveals that residual errors of the VGGT classifier are \emph{highly structured} and dominated by a few recurring confusion patterns rather than broadly distributed noise.

First, there is a pronounced tendency to over-predict \texttt{\small dolly out} as a false positive across multiple ground-truth classes: both \texttt{\small crane up} and \texttt{\small crane down} are frequently mislabeled as \texttt{\small dolly out} (\eg, 15 and 9 counts, respectively), and similarly for \texttt{\small truck left/right} (11 counts each) and \texttt{\small tilt up/down} (8--7 counts). This suggests that, in some segments, outward depth change acts as an ``attractor" class, likely because multiple camera trajectories contain a depth component and the mapping reduces a continuous pose trajectory to discrete primitives.

Second, we observe systematic confusions among coupled motions and axis-aligned primitives. In particular, \texttt{\small arc cw/ccw} is often confused with \texttt{\small dolly out} and with horizontal motion primitives (\eg, \texttt{\small pan right}), consistent with the fact that \texttt{\small arc} combines yaw rotation with lateral translation and may also include a depth component depending on the trajectory. For example, \texttt{\small arc cw} is frequently predicted as \texttt{\small dolly out} or \texttt{\small pan right}, while \texttt{\small arc ccw} shows confusion with \texttt{\small truck right} and \texttt{\small dolly out}.

Third, the largest single off-diagonal entry corresponds to \texttt{\small dolly in} being misclassified as \texttt{\small pan right} (16 counts), indicating that certain trajectories exhibit concurrent yaw change and forward motion where the dominant primitive is ambiguous at 1-second granularity. Similar interactions appear between \texttt{\small tilt} and translation primitives, where pitch changes co-occur with camera motion and lead to mixed predictions.

Finally, \texttt{\small static} errors are sparse but concentrated: when misclassified, \texttt{\small static} is most often predicted as \texttt{\small dolly out} (3 counts), consistent with threshold-boundary cases where small residual translation exceeds the ``no dominant motion'' criterion.

Overall, the confusion matrix indicates that the remaining failure modes are driven primarily by (i) discretization of continuous camera trajectories into primitives and (ii) co-occurring multi-axis motions where one component (notably depth change) is over-selected. This motivates both the constrained multi-label formulation (to allow co-occurrence) and future improvements in pose-to-label mapping (\eg, better dominance criteria for separating depth change from vertical/lateral translation in short clips).

\section{Full Prompts and additional Qualitative Examples}

\subsection{Full prompt templates}
\label{sec:supp_prompts}

We evaluate how explicit camera-motion cues affect \emph{VideoLLM} generation using three prompts that share the same video input (a sequence of uniformly sampled frames) and differ only in whether and how motion information is communicated. The \emph{motion-header prompt} follows the structured prompting design in the main paper (Sec.~3.5), where a per-second motion list is prepended as a compact textual scaffold~\cite{lin2025towards}.

\noindent\textbf{Baseline prompt (generic description).}
This is the default VideoLLM usage pattern: the model receives frames and a generic captioning instruction, without any explicit cinematography cues.
\begin{quote}
\ttfamily\footnotesize
Here are [N] consecutive video frames. They are evenly sampled at a frame rate of [r] FPS.

Describe the video clip using clear and concise language. Make your description in one paragraph.
\end{quote}
In practice, this prompt often yields content-correct but camera-vague descriptions, and may conflate camera motion with object motion.

\noindent\textbf{Structured prompt (filmmaker-style instruction).}
Motivated by the qualitative prompting protocol, we reformulate the instruction in filmmaking language to explicitly request cinematographic attributes (lighting, framing, composition) and to encourage temporal linking across frames.
\begin{quote}
\ttfamily\footnotesize
Here are [N] consecutive video frames. They are evenly sampled at a frame rate of [r] FPS.

Describe this video using the filmmaker's language, highlighting the lighting, framing, video composition, and especially camera usage that connects different frames. For example: ``At the beginning, <video content>; then <camera motion>, <video content>; ...; finally, <camera motion>, <video content>''. Make your description in a paragraph.
\end{quote}
This structure improves the \emph{intent} to discuss camera usage, but it still requires the VideoLLM to infer motion direction and temporal evolution solely from its visual representation.

\noindent\textbf{Camera motion injected structured prompt.}
We inject predicted motion primitives as a short header before the same filmmaker-style instruction. Following Sec.~3.5 of the main paper, a shot with $S$ one-second segments is serialized as a list
\texttt{\small Per-second camera motion: [m$_1$, ..., m$_S$]}, where each $m_s$ is a canonicalized set of primitives (\eg, \texttt{\small static} or \texttt{\small pan left + tilt up}).
\begin{quote}
\ttfamily\footnotesize
Here are [N] consecutive video frames. They are evenly sampled at a frame rate of [r] FPS.

Per-second camera motion: [m1, m2, ..., mS].

Describe this video using the filmmaker's language, highlighting the lighting, framing, video composition, and especially camera usage that connects different frames. For example: ``At the beginning, <video content>; then <camera motion>, <video content>; ...; finally, <camera motion>, <video content>''. Make your description in a paragraph.
\end{quote}
This modification is \emph{training-free}: it does not change VideoLLM weights or architecture, but conditions generation on an explicit motion prior.

We design the motion header to be \emph{compact}, \emph{structured}, and \emph{physically grounded} (primitive taxonomy with constraints). This choice is motivated by two empirical observations. First, probing shows that camera motion is only weakly recoverable from frozen VideoLLM vision features, making direction and temporal consistency brittle without additional cues. Second, camera motion is intrinsically compositional and time-indexed; a per-second list matches the segment granularity of our benchmark and provides a stable alignment target for the language model. As discussed in the main paper, providing explicit per-second motion labels encourages temporally grounded narratives, reduces motion-direction hallucinations, and biases the model toward continuity-aware reasoning (\eg, linking subject changes to camera transitions) rather than producing generic cinematography statements.

\subsection{More qualitative examples}
\label{sec:supp_qual}

To further illustrate the effect of camera-motion prompting, we provide additional qualitative examples from \emph{CineTechBench}~\cite{wang2025cinetechbench}. 
For each clip, we compare the outputs produced by the three prompt variants described in Sec.~\ref{sec:supp_prompts}: 
\emph{baseline}, \emph{structured}, and our \emph{motion-header} prompt.

\begin{figure*}[t]
\centering
\includegraphics[width=\textwidth]{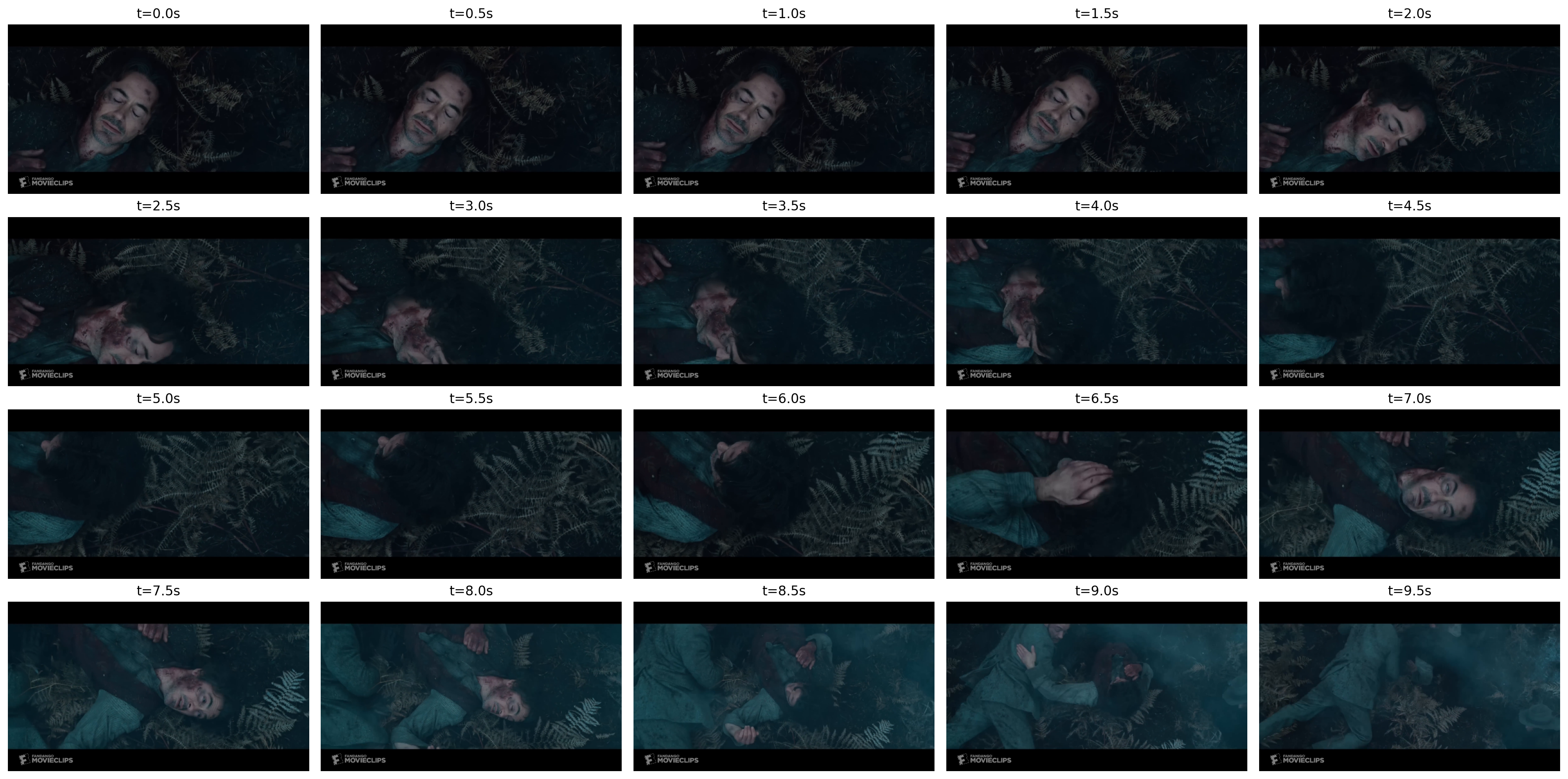}
\caption{Clip2 sampled frames from CineTechBench~\cite{wang2025cinetechbench}.}
\label{fig:supp_clipA}
\end{figure*}

\begin{figure*}[t]
\centering
\includegraphics[width=\textwidth]{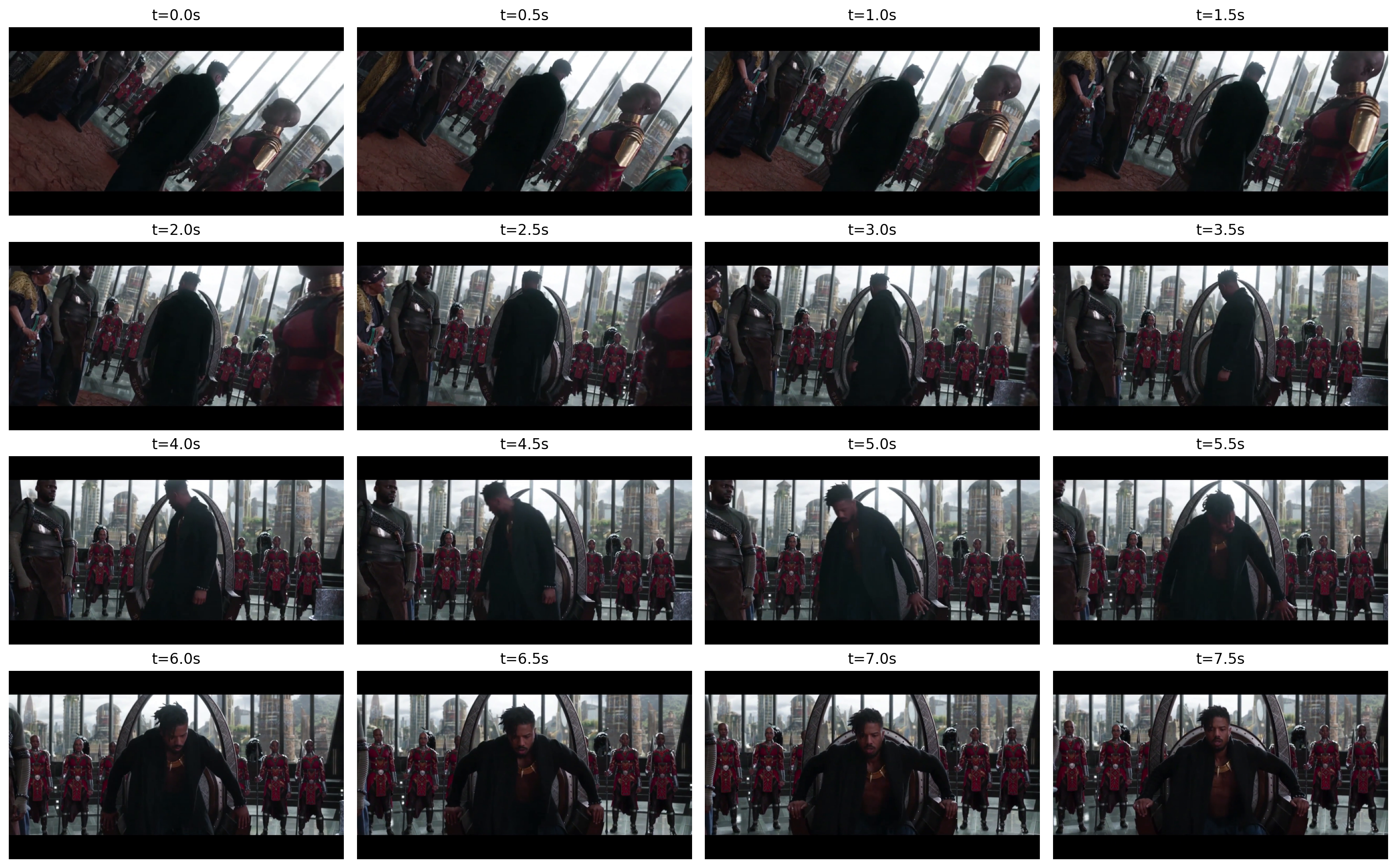}
\caption{Clip3 sampled frames from CineTechBench~\cite{wang2025cinetechbench}.}
\label{fig:supp_clipB}
\end{figure*}

Figures~\ref{fig:supp_clipA} (Clip2) and~\ref{fig:supp_clipB} (Clip3) show sampled frames from two example clips. 
Together with the example shown in the main paper (Clip1), we present three representative cases covering different camera motions and scene types. For each clip, we report the generated descriptions under the three prompt settings, resulting in nine outputs in total (\cref{tab:supp_qual_results},~\ref{tab:supp_qual_results2},~\ref{tab:supp_qual_results3}). 
Key phrases related to camera motion are highlighted for comparison.

\begin{table*}[t]
\centering
\footnotesize
\setlength{\tabcolsep}{3pt}
\renewcommand{\arraystretch}{1.0}
\caption{Qualitative comparison of model-generated descriptions for Clip1 under different prompt templates. Camera motion-related phrases are highlighted in bold.}
\label{tab:supp_qual_results}
\begin{tabular}{p{5.0cm}|p{6.0cm}|p{6.0cm}}
\toprule
\textbf{Baseline prompt} & \textbf{Structured prompt (filmmaker-style instruction)} & \textbf{Camera motion injected structured prompt} \\
\midrule
This video portrays a dramatic musical performance scene. It begins with a conductor, dressed in black, gesturing with precision and intensity, likely cueing the start of a musical piece. His facial expression is focused and assertive, signaling a moment of high tension. The camera quickly \textbf{pans} with \textbf{motion blur}, transitioning to a drummer mid-performance, energetically striking the drums while accompanied by a bassist. The \textbf{motion blur} appears again between frame changes, emphasizing swift \textbf{camera movement} or \textbf{cuts} to maintain rhythmic pacing. The conductor reappears in subsequent frames, continuing to direct with expressive hand movements, culminating in another quick \textbf{blur} before returning to the drummer, now deeply immersed in an intense and possibly climactic moment. The alternating shots and dynamic movement suggest a tightly \textbf{edited sequence} synchronized with musical tempo, evoking a sense of urgency and artistic fervor typical of a performance climax.
&
At the beginning, the video opens with a medium close-up of the conductor, dramatically lit with focused, warm stage lighting that casts soft shadows and isolates him against the dimly lit auditorium, emphasizing his stern facial expression and sharp, deliberate hand gestures. Then, a rapid \textbf{\color{VioletRed}whip pan to the right} (evident from \textbf{motion blur}) bridges the visual transition, dynamically redirecting the viewer's focus to the drummer. The \textbf{camera settles} into a medium-wide shot of the drum set, capturing the drummer mid-performance under similarly warm, directional lighting that accentuates the shimmer of the cymbals and the energy of his movements. The \textbf{framing} includes part of the upright bassist, preserving ensemble context while maintaining the drummer as the focal point. Another \textbf{whip pan} follows, visually echoing the earlier motion and swiftly returning the viewer to the conductor. As the video \textbf{cuts back}, the conductor appears more animated in his movements, with tight \textbf{framing} and consistent lighting maintaining visual continuity. A final \textbf{whip pan} transitions back to the drummer, who now performs with increased intensity, reinforcing the escalating rhythm and emotional tension through repeated visual interplay. The sequence's \textbf{camera work}--marked by fluid \textbf{whip pans}--mirrors the musical tempo, while the lighting and composition maintain dramatic cohesion, heightening the confrontation and synchrony between conductor and performer.
&
At the beginning, under warm, low-key stage lighting, a \textbf{static} medium closeup \textbf{frames} the conductor from stage right, isolating him against a softly lit auditorium background as he cues the ensemble with precise hand gestures. Then, a \textbf{pan-left} motion bridges the space, transitioning fluidly into a medium shot of the drummer, dynamically lit from above to emphasize the percussive energy and \textbf{motion blur} of his performance. Next, a \textbf{pan-right} swiftly returns the focus to the conductor, capturing his sharp directional cues in a near-symmetrical composition that mirrors the earlier frame, reinforcing continuity through motion. A \textbf{static shot} then \textbf{holds} on the conductor, now mid-gesture, in tighter \textbf{framing} that heightens dramatic tension while maintaining spatial coherence. Finally, a \textbf{pan-left} reintroduces the drummer in a more intensified moment--his kinetic energy visually amplified through rhythmic lighting and a centered composition that anchors the visual climax. The sequence skillfully employs lateral \textbf{camera movement} and repeated \textbf{framing} motifs to synchronize the visual rhythm with the musical performance, creating a cohesive and propulsive cinematic language.
\\

\bottomrule
\end{tabular}
\end{table*}

\begin{table*}[t]
\centering
\footnotesize
\setlength{\tabcolsep}{3pt}
\renewcommand{\arraystretch}{1.0}
\caption{Qualitative comparison of model-generated descriptions for Clip2 under different prompt templates. Camera motion-related phrases are highlighted in bold.}
\label{tab:supp_qual_results2}
\begin{tabular}{p{5.0cm}|p{6.0cm}|p{6.0cm}}
\toprule
\textbf{Baseline prompt} & \textbf{Structured prompt (filmmaker-style instruction)} & \textbf{Camera motion injected structured prompt} \\
\midrule
The clip shows a wounded man lying motionless on the forest floor among ferns and leaves, his face bruised and smeared with blood, suggesting he has been injured or unconscious. The \textbf{camera gradually shifts its framing} across the scene, revealing more of his body and the surrounding foliage as his head slowly tilts to the side. After remaining still for several moments, he begins to stir slightly, moving his head and hand as if regaining consciousness. Toward the end of the clip, another person enters from the left side, reaching toward him and leaning over his body, apparently checking on him or attempting to help him while the injured man weakly reacts.
&
At the beginning, the shot opens in a dim, cool-toned forest, lit by soft bluish ambient light. A tight overhead close-up \textbf{frames} a wounded man lying among ferns and damp leaves, his face near the center while foliage fills the surrounding space. As the sequence progresses, the \textbf{camera slowly drifts laterally while rolling}, shifting the man's face from a frontal view to a slightly oblique angle. The \textbf{framing gradually widens} from a centered close-up to a looser, off-center composition, revealing more of his shoulder and the forest floor. The \textbf{\color{VioletRed}continuous roll with gentle tracking} slides his head toward the lower part of the frame while foliage expands across the upper area, creating a drifting, hovering sensation. Toward the end, the \textbf{camera dollies slightly outward while maintaining the roll}, widening the overhead view as another person enters from the left and leans toward the injured man. The shot settles into a \textbf{loose overhead two-shot}, both figures partially visible among the ferns, with the slow rolling camera motion unifying the sequence into a calm but tense visual moment.
&
The shot opens in a dim, cool-toned forest with soft bluish lighting. A wounded man lies among ferns and damp leaves, \textbf{framed in a tight overhead close-up} that centers on his face and upper torso. The camera \textbf{rolls slowly clockwise}, keeping the close framing while rotating the horizon so the foliage appears to drift around him, creating a disorienting sense of instability. Midway through, the motion becomes more complex: the camera continues the \textbf{clockwise roll} while \textbf{trucking left} and \textbf{craning up}, gradually widening the view and revealing more of the forest floor. Soon after, the camera maintains the roll and begins a gentle \textbf{dolly out}, expanding the overhead composition. As the frame widens, another figure enters from the left, kneels beside the injured man, and reaches toward him. The shot ends in a wider overhead view, with both figures visible among the ferns as the slow \textbf{rolling, outward-moving camera} connects the sequence into a somber, atmospheric moment.
\\
\bottomrule
\end{tabular}
\end{table*}

\begin{table*}[t]
\centering
\footnotesize
\setlength{\tabcolsep}{3pt}
\renewcommand{\arraystretch}{1.0}
\caption{Qualitative comparison of model-generated descriptions for Clip3 under different prompt templates. Camera motion-related phrases are highlighted in bold.}
\label{tab:supp_qual_results3}
\begin{tabular}{p{5.0cm}|p{6.0cm}|p{6.0cm}}
\toprule
\textbf{Baseline prompt} & \textbf{Structured prompt (filmmaker-style instruction)} & \textbf{Camera motion injected structured prompt} \\
\midrule
A man wearing a long dark coat walks forward through a grand, modern throne room with floor-to-ceiling windows overlooking a futuristic city. Several guards in coordinated red uniforms stand symmetrically on both sides of the room, holding spears and watching as he approaches a large, curved ceremonial throne positioned at the center. Other armored attendants and officials stand nearby, observing quietly. The man slows as he reaches the throne, turns slightly, then lowers himself into the seat, gripping the armrests and settling into a commanding posture while the guards remain lined up behind him, reinforcing the formal and authoritative atmosphere of the scene.
&
At the beginning, a slightly \textbf{canted} medium-wide establishing shot inside a grand, glass-walled hall, where bright natural daylight floods in from the floor-to-ceiling windows behind the characters, producing strong backlighting and soft silhouettes. In the foreground, a man in a long dark coat walks toward a large circular throne-like structure positioned at the center of the frame, while symmetrical rows of red-armored guards stand rigidly in the background, forming a visual corridor that guides the viewer's eye inward. As the shot progresses, the \textbf{\color{VioletRed}camera slowly tracks forward} behind him, gradually straightening its angle and \textbf{tightening the framing}, turning the initial off-axis composition into a more centered, ceremonial perspective. The circular throne becomes a dominant compositional element, \textbf{framing} the character as he approaches; the guards remain evenly spaced on both sides, reinforcing the symmetry and depth of the mise-en-scène. Continuing the forward \textbf{push-in}, the camera subtly \textbf{lowers and stabilizes into a frontal alignment} as the man reaches the throne, placing him precisely on the central axis of the architecture and the guard formation. He then turns and lowers himself into the seat, the movement captured in a steady medium shot that emphasizes the ritualistic nature of the action. Finally, as he settles into the throne and grips the armrests, the camera completes its \textbf{\color{VioletRed}push-in} to a centered, balanced composition: the character sits framed by the curved throne behind him, flanked symmetrically by the guards and the luminous cityscape beyond the windows, creating a powerful throne-room tableau under strong backlighting.
&
At the beginning, a slightly \textbf{canted wide shot} reveals a grand, glass-walled hall flooded with soft daylight from floor-to-ceiling windows, the bright skyline forming a luminous backdrop while a group of red-armored guards stands in symmetrical formation. A man in a dark coat walks toward a sculptural throne framed by two curved metallic blades. As the camera \textbf{dollies in while rolling clockwise}, the off-axis framing creates a subtle sense of motion and tension, gradually tightening the composition around the man's back and the throne. The \textbf{dolly-in with clockwise roll continues}, bringing the throne into clearer alignment behind him while the guards remain evenly spaced on both sides, their red armor catching rim light from the windows. The lighting stays high-key and backlit, outlining figures with soft highlights while the foreground remains slightly shadowed, adding depth. As the camera \textbf{keeps dollying in and rolling clockwise}, the frame slowly levels and the throne becomes centered in the composition. The man turns and lowers himself toward the seat, the curved blades of the throne visually framing his body. Finally, the \textbf{dolly-in with clockwise roll resolves into a centered medium shot}, where he settles into the throne and grips the armrests, with the guards standing symmetrically behind him and the bright cityscape still glowing through the windows, completing the camera's smooth approach and stabilizing the composition around his newly assumed position of authority.
\\
\bottomrule
\end{tabular}
\end{table*}

Across the three examples, the baseline prompt primarily focuses on scene content and rarely provides reliable descriptions of camera motion. The structured prompt encourages the model to discuss cinematographic aspects, but the inferred motion cues are often ambiguous or incorrect. For instance, in Clip1, the structured prompt incorrectly describes the motion as a \emph{whip pan to the right}, while the injected motion header leads the model to correctly describe alternating \emph{pan-left} and \emph{pan-right} movements that match the camera dynamics. In Clip2, the structured prompt identifies a rolling camera but does not clearly specify its direction or temporal progression, whereas the motion-header prompt explicitly grounds the description in a \emph{clockwise roll} combined with \emph{trucking} and \emph{dolly-out} movements, producing a more precise and temporally coherent narrative. In Clip3, the structured prompt relies on loosely defined cinematographic terms such as ``push-in'' or ``tracking forward,'' which describe similar motion but lack a consistent taxonomy and omit the camera roll entirely. In contrast, the motion-header prompt enforces explicit primitives such as \emph{dolly-in} and \emph{roll clockwise}, yielding a more unified and geometrically grounded description. These examples highlight that while structured prompting alone encourages camera-aware language, providing explicit motion primitives substantially improves directional correctness, terminological consistency, and temporal grounding in VideoLLM-generated descriptions.

\section{Additional Discussion}
\label{sec:supp_discussion}

\subsection{Synthetic-to-real gap}
\label{sec:supp_syn2real}

Our benchmark is deliberately built on synthetic data from MultiCamVideo~\cite{bai2025recammaster} to obtain ground-truth camera extrinsic parameters that are unavailable in real-world datasets. This design choice provides precise, physically grounded labels free from annotation noise, but introduces a domain gap with respect to real-world video characteristics such as lens distortion, motion blur, rolling shutter artifacts, and non-uniform lighting.

We view this synthetic benchmark as a \emph{diagnostic tool} rather than a claim of real-world coverage: it isolates camera motion understanding from confounds that plague manual annotation (\eg, visual illusions caused by object motion, subjective labeling of ambiguous clips). Limited by time and computational resources, evaluation and exploration of real-world applications are not covered in this work. To mitigate the gap, several strategies can be pursued in future work:
\begin{itemize}
    \item \textbf{Real-world evaluation.} Benchmarks such as CameraBench~\cite{lin2025towards} provide human-annotated camera motion labels for real videos, albeit at coarser granularity and with camera motions not fully defined in the camera coordinate system. Testing our pipeline on CameraBench would quantify how well the VGGT-based classifier generalizes beyond the synthetic domain. We note that VGGT itself is pretrained on diverse real-world data, which may facilitate transfer.
    \item \textbf{Domain adaptation via augmentation.} We outline a data augmentation strategy (Sec.~\ref{sec:supp_augmentation}) that introduces realistic visual effects (zoom, motion blur, camera shake) to narrow the appearance gap.
    \item \textbf{Mixed training.} Future work can combine our synthetic labels with pseudo-labels from real videos (\eg, generated by running SfM on in-the-wild footage) to improve robustness.
\end{itemize}

We emphasize that the primary contribution of this work is to \emph{inspire the VLM community} to consider explicit embedding or incorporation of camera motion information. The synthetic benchmark serves as a proof of concept demonstrating: (1) current VideoLLMs have a measurable camera-motion blindness; (2) geometric cues from 3DFMs can substantially improve recognition; and (3) structured prompting is a viable injection mechanism. These findings are independent of the data domain and motivate further investigation on real-world data.

\subsection{Camera intrinsics and zoom}
\label{sec:supp_zoom}

Our current taxonomy covers 15 primitives derived entirely from camera extrinsic parameters (translation and rotation), intentionally omitting intrinsic changes such as zoom (focal-length variation). This is a notable limitation, as zoom is a fundamental cinematographic technique that is frequently confused with dolly motion~\cite{lin2025towards}.

The omission is a deliberate scope choice: MultiCamVideo provides precise extrinsic trajectories but does not vary focal length across frames, making it unsuitable for zoom annotation. Extending the taxonomy to include zoom requires either:
\begin{itemize}
    \item \textbf{Synthetic data with intrinsic variation.} Rendering engines can simulate zoom by programmatically varying the virtual camera's focal length. Combined with the existing extrinsic trajectories, this would yield a richer motion space. We outline a zoom simulation strategy via post-hoc cropping in Sec.~\ref{sec:supp_augmentation}.
    \item \textbf{Real-world intrinsic estimation.} Modern 3DFMs and SfM pipelines can jointly estimate camera intrinsics from video; however, disentangling true optical zoom from digital zoom or dolly-induced scale change remains challenging.
\end{itemize}

We note that our pipeline architecture is agnostic to the number of motion primitives: the multi-label classifier and incompatibility matrix can be extended to include zoom-in and zoom-out as additional labels once appropriate training data is available. In the data augmentation section (Sec.~\ref{sec:supp_augmentation}), we propose a series of augmentations that cover zoom-in and zoom-out simulation (while not reflecting true optical changes). Training on the augmented data does not require any architectural change and can be a practical step toward improving robustness to zoom-like effects in real videos.

\subsection{Joint training vs.\ plug-and-play}
\label{sec:supp_joint_training}

A natural question is whether directly feeding VGGT features into a VideoLLM and jointly training on camera motion data would yield better results than our plug-and-play approach. We chose the latter for several reasons:

\begin{itemize}
    \item \textbf{Model agnosticism.} Our pipeline works with any VideoLLM without modifying its weights. Joint training would require per-model adaptation and access to the original training infrastructure.
    \item \textbf{No VideoLLM fine-tuning cost.} Fine-tuning a 7B+ VideoLLM on camera motion data risks catastrophic forgetting of general video understanding capabilities, requiring careful multi-task training.
    \item \textbf{Modularity and interpretability.} The structured prompt provides an explicit, human-readable interface between camera cues and the VideoLLM, making it straightforward to verify, debug, or override motion predictions.
\end{itemize}

Recent work such as VLM-3R~\cite{fan2025vlm3r} demonstrates that trainable 3D-enhanced VLMs can achieve strong spatial-temporal understanding by fusing camera tokens during training. We view our plug-and-play approach and trainable approaches as complementary: ours provides an immediate, low-cost solution, while joint training may achieve higher ceilings given sufficient compute and data. Limited by computational resources, the joint-training approach is not adopted and evaluated in this work, while a systematic comparison between these paradigms is an important direction for future work.

\subsection{VGGT dependency and practical deployment}
\label{sec:supp_vggt_practicality}

The strongest variant of our pipeline relies on VGGT (1.2B parameters), which adds computational overhead. We address this concern through three observations:

\begin{itemize}
    \item \textbf{Distillation reduces cost substantially.} The VGGT--Q-Former distillation achieves 5.3$\times$ throughput improvement at 39\% peak memory, with only a 10-point drop in instance-level accuracy. This demonstrates distillation as a promising path to making the pipeline practical for applications that require real-time or large-scale inference.
    \item \textbf{The 3DFM ecosystem is rapidly improving.} LiteVGGT~\cite{shu2025litevggt} already reduces VGGT's cost through geometry-aware token merging. As lighter 3DFMs emerge, our pipeline directly benefits without architectural changes. Moreover, camera awareness can complement the VLM and VLA communities' ongoing efforts to incorporate 3D geometry into video understanding.
    \item \textbf{Camera cues are computed offline.} For applications like media indexing, retrieval, or batch captioning, VGGT inference can be amortized as a one-time preprocessing step, after which only the lightweight classifier runs online.
\end{itemize}

\section{Proposed Data Augmentation Strategy}
\label{sec:supp_augmentation}

\begin{figure}[t]
\centering
\includegraphics[width=\columnwidth]{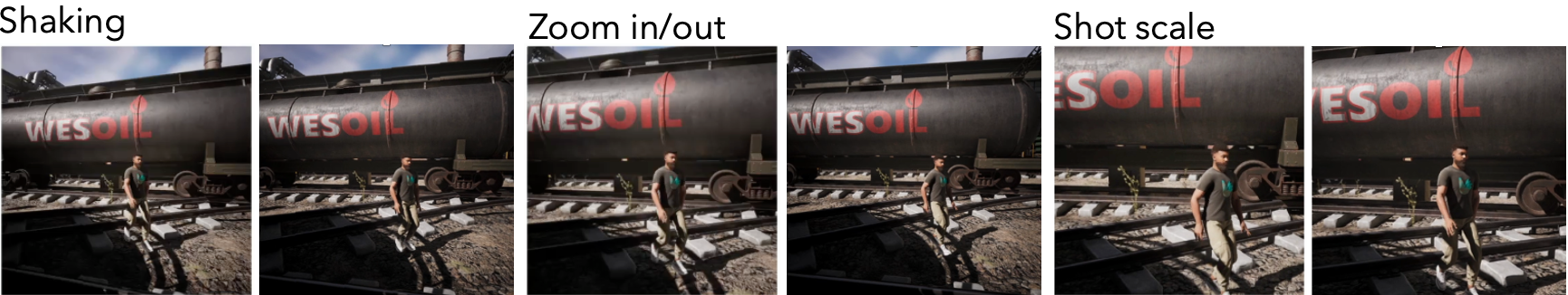}
\caption{A visualization of the proposed data augmentations. For each augmentation, the left column shows the starting frames and the right column shows the augmented ending frames, because the augmentation is applied to the entire segment.}
\label{fig:supp_aug}
\end{figure}

To improve robustness and bridge the synthetic-to-real gap, we propose the following augmentation strategies applicable during classifier training. Each augmentation must respect compatibility with existing camera motion labels (Fig.~\ref{fig:supp_aug}).

\begin{enumerate}
    \item \textbf{Zoom-in / Zoom-out simulation.} Since the source data lacks focal-length variation, we simulate zoom effects via progressive center cropping (zoom-in) or padding with border extrapolation (zoom-out) across frames in a segment. The crop ratio is linearly interpolated from 1.0 to a target ratio sampled in $[0.6, 0.9]$ for zoom-in, and reversed for zoom-out. \emph{Compatibility:} zoom-in and dolly-in should not co-occur as labels; an updated incompatibility matrix enforces this constraint.
    \item \textbf{Reverse play.} Temporally reversing a segment swaps directional labels (\eg, \texttt{\small pan-left} $\leftrightarrow$ \texttt{\small pan-right}, \texttt{\small dolly-in} $\leftrightarrow$ \texttt{\small dolly-out}), effectively doubling the training set with consistent labels. \emph{Compatibility:} all directional primitives must be symmetrically swapped.
    \item \textbf{Random center cropping.} Cropping with a ratio sampled in $[0.6, 0.9]$ simulates varying fields of view and removes border artifacts. Since the crop is spatially centered and consistent across frames, it does not alter camera motion labels. \emph{Compatibility:} compatible with all motion primitives.
\end{enumerate}

These augmentations are designed to be applied \emph{jointly} with the original camera motion labels, requiring no re-annotation. The augmentation pipeline can be integrated into the training dataloader with minimal overhead. We also considered \textbf{camera shake simulation} (per-frame random translations and roll rotations with temporal smoothing) and \textbf{motion blur via frame blending} ($\alpha$-compositing of adjacent frames). However, we did not implement these two augmentations: the degree of shake and blur are difficult to label in a categorical framework, and more importantly, detecting their presence in real footage is efficiently handled by conventional signal-processing methods, making model-based prediction less beneficial for these attributes compared to semantic camera motion primitives.

\section{LLM-as-Judge Evaluation Protocol}
\label{sec:supp_llm_judge}

To enable systematic evaluation of camera-aware video descriptions beyond qualitative inspection, we propose an LLM-as-judge protocol~\cite{zheng2023judging}. This protocol uses a large language model to score VideoLLM-generated descriptions along multiple cinematographic dimensions. Fig.~\ref{fig:supp_llm_judge} provides the full prompt template, which includes detailed scoring criteria for each dimension. The LLM evaluates each description based on its alignment with ground-truth camera motion labels, temporal coherence, reproducibility of camera work, narrative tone, and language quality. This approach enables scalable, consistent evaluation that complements human judgment and provides quantitative metrics for future research.

Each generated description is evaluated on five dimensions, each scored from 1 (poor) to 5 (excellent):

\begin{enumerate}
    \item \textbf{Cinematographic Accuracy (CA).} Does the description correctly identify camera motion types, shot scales, and framing? Are technical terms (pan, tilt, dolly, truck, crane, static) used accurately and consistently with the ground-truth motion labels?

    \item \textbf{Temporal Continuity (TC).} Does the description follow the temporal progression of the video? Are camera motions described in the correct chronological order? Does it capture transitions between different motion phases?

    \item \textbf{Reproducibility Detail (RD).} Could a cinematographer recreate the described camera work from the description alone? Are directions, durations, and spatial relationships specified precisely enough?

    \item \textbf{Narrative Mood \& Tone (NM).} Does the description convey the artistic intent and emotional quality of the camera work? Does it connect camera choices to narrative effect?

    \item \textbf{Language Quality (LQ).} Is the description fluent, well-structured, and free of contradictions? Does it use professional cinematographic vocabulary appropriately?
\end{enumerate}

\begin{figure*}[t]
\begin{tcolorbox}[colback=blue!5!white, colframe=blue!40!white, title={\textbf{LLM-as-Judge Prompt Template}}, fonttitle=\small\bfseries, coltitle=black, boxrule=0.5pt, arc=2pt]
\footnotesize\ttfamily
You are an expert cinematographer evaluating a video description. You are given:\\[2pt]
- Ground-truth per-second camera motion labels: [\{motion\_labels\}]\\
- The generated video description: [\{description\}]\\[4pt]
Score the description from 1 to 5 on each of the following dimensions:\\[4pt]
\textbf{1. Cinematographic Accuracy (CA)}\\
Assess whether the description correctly and precisely identifies camera angle, framing, lens focal length, camera movement (direction and speed), lighting type and direction, and composition.\\
\textrm{\textbullet}~5: All key elements correct, using industry-standard terminology.\\
\textrm{\textbullet}~4: Mostly accurate with minor omissions or ambiguity.\\
\textrm{\textbullet}~3: Some inaccuracies or generalizations; partially correct terminology.\\
\textrm{\textbullet}~2: Several incorrect elements; relies on non-technical language.\\
\textrm{\textbullet}~1: Inaccurate or vague; lacks meaningful cinematographic detail.\\[4pt]
\textbf{2. Temporal Continuity (TC)}\\
Measure how clearly the description communicates the sequence of events, camera transitions, and rhythmic or pacing logic throughout the scene.\\
\textrm{\textbullet}~5: Seamless, accurate timeline; per-shot motion and transitions are logically and temporally clear.\\
\textrm{\textbullet}~4: Mostly clear sequence with minor inconsistencies or vague transitions.\\
\textrm{\textbullet}~3: Partial timeline or implied transitions; lacks precise continuity.\\
\textrm{\textbullet}~2: Confusing or incorrect sequence; missing or inaccurate motion description.\\
\textrm{\textbullet}~1: Lacks chronological structure; incoherent scene progression.\\[4pt]
\textbf{3. Reproducibility Detail (RD)}\\
Evaluate whether the description provides sufficient technical detail for a filmmaker to recreate the scene, including shot scale, camera positioning, blocking, movement mechanics, lighting setup, frame duration, and scene geometry.\\
\textrm{\textbullet}~5: Fully reproducible; contains all necessary technical elements.\\
\textrm{\textbullet}~4: Mostly reproducible; a few minor missing parameters.\\
\textrm{\textbullet}~3: Some useful information; key elements (e.g., camera height, lighting angles) are missing.\\
\textrm{\textbullet}~2: Vague and incomplete; only general scene description present.\\
\textrm{\textbullet}~1: Not reproducible; lacks all relevant production-level detail.\\[4pt]
\textbf{4. Narrative Mood \& Tone (NM)}\\
Measure the description's ability to convey the emotional atmosphere, dramatic intention, and cinematic tone (e.g., tension, melancholy, exuberance), as well as the subjective energy of the scene.\\
\textrm{\textbullet}~5: Strong emotional clarity aligned with visual style and editing rhythm.\\
\textrm{\textbullet}~4: Clear tone with some expressive nuance.\\
\textrm{\textbullet}~3: Functional tone description; lacks depth or resonance.\\
\textrm{\textbullet}~2: Tone is underdeveloped or inconsistent.\\
\textrm{\textbullet}~1: No discernible mood or emotional quality.\\[4pt]
\textbf{5. Language Quality \& Fluency (LQ)}\\
Assess overall clarity, precision, and fluency of the writing, with a focus on cinematographic vocabulary, cohesion, and technical correctness in grammar and syntax.\\
\textrm{\textbullet}~5: Polished, technically fluent, and lexically precise.\\
\textrm{\textbullet}~4: Clear and correct with minor stylistic or lexical weaknesses.\\
\textrm{\textbullet}~3: Mostly clear; occasional awkward phrasing or imprecise terminology.\\
\textrm{\textbullet}~2: Grammatically flawed or inconsistent in tone or register.\\
\textrm{\textbullet}~1: Unclear, ungrammatical, or non-professional.\\[4pt]
For each dimension, provide: (1) Score (1--5), and (2) One-sentence justification.\\[2pt]
Finally, compute the weighted average:\\
Final = 0.30*CA + 0.25*TC + 0.25*RD + 0.10*NM + 0.10*LQ
\end{tcolorbox}
\caption{The full prompt template for the LLM-as-judge evaluation protocol.}\label{fig:supp_llm_judge}
\end{figure*}

\subsection{Preliminary Results}

To validate the evaluation protocol, we apply it to descriptions generated by GPT-4o from a single example clip (Fig. 7 in the main text). We compare the three prompting strategies used in the qualitative analysis:
\begin{enumerate}
    \item[\textbf{P1.}] \emph{Baseline}: ``Describe this video in detail in a paragraph.''
    \item[\textbf{P2.}] \emph{Filmmaker's language}: Request a description using cinematographic terminology (lighting, framing, composition, camera usage) with a temporal structure template.
    \item[\textbf{P3.}] \emph{Filmmaker's language + motion labels}: Same as P2, with ground-truth per-second camera motion labels prepended (\eg, \texttt{\small [static, pan-left, pan-right, static, pan-left]}).
\end{enumerate}

\begin{table}[t]
\centering
\small
\setlength{\tabcolsep}{3pt}
\renewcommand{\arraystretch}{1.1}
\caption{\textbf{LLM-as-judge scores for three prompting strategies.} Scores range from 1 (poor) to 5 (excellent). Injecting camera motion labels (P3) yields the largest gains in cinematographic accuracy and temporal continuity.}
\label{tab:supp_judge_scores}\vspace{-0.2cm}
\begin{tabular}{lcccccc}
\toprule
\textbf{Prompt} & \textbf{CA} & \textbf{TC} & \textbf{RD} & \textbf{NM} & \textbf{LQ} & \textbf{Final} \\
\midrule
P1: Baseline           & 1 & 2 & 1 & 3 & 4 & 1.75 \\
P2: Filmmaker         & 3 & 3 & 4 & 5 & 5 & 3.65 \\
P3: + Motion labels   & 4 & 4 & 3 & 4 & 5 & 3.85 \\
\bottomrule
\end{tabular}\vspace{-0.2cm}
\end{table}

Several observations emerge. The baseline prompt (P1) produces fluent but cinematographically vague descriptions, scoring lowest on accuracy and reproducibility. Requesting filmmaker's language (P2) substantially improves reproducibility and narrative tone, but camera motion directions remain incorrect (\eg, describing a ``whip pan to the right'' when ground truth is pan-left), limiting the accuracy and temporal continuity scores. Injecting ground-truth motion labels (P3) corrects these directional errors and improves temporal alignment, raising the final score from 3.65 to 3.85. The largest per-dimension gain appears in cinematographic accuracy (+1) and temporal continuity (+1), directly validating the core hypothesis of our structured prompting approach: providing explicit camera motion cues enables more precise and faithful video descriptions.

These preliminary results, while based on a single example, demonstrate that the evaluation protocol can meaningfully discriminate between prompting strategies of varying informativeness, and that the structured motion header is the key factor driving improvement. We leave large-scale evaluation across the full dataset to future work.

{
    \small
    \bibliographystyle{ieeenat_fullname}
    \bibliography{main}
}

%% file: preamble.tex
\usepackage[normalem]{ulem}

\definecolor{mypurple}{rgb}{0.8, 0.3764705882352941, 1.0}

%% file: sec/0_abstract.tex
\begin{abstract}
Camera motion is a fundamental geometric signal that shapes visual perception and cinematic style, yet current video-capable vision-language models (VideoLLMs) rarely represent it explicitly and often fail on fine-grained motion primitives. We address this gap with a framework of \textbf{benchmarking}, \textbf{diagnosis}, and \textbf{injection}. We derive \textbf{CameraMotionDataset}, a VQA benchmark built on an existing synthetic dataset (MultiCamVideo Dataset) with explicit camera control, formulate camera motion as constraint-aware multi-label recognition, and construct a multiple-choice evaluation protocol--\textbf{CameraMotionVQA}. Across diverse off-the-shelf VideoLLMs, we observe substantial errors in recognizing camera motion primitives. Probing experiments on a Qwen2.5-VL vision encoder suggest that camera motion cues are weakly represented, especially in deeper ViT blocks, helping explain the observed failure modes. To bridge this gap without costly training or fine-tuning, we propose a lightweight, model-agnostic pipeline that extracts geometric camera cues from 3D foundation models (3DFMs), predicts constrained motion primitives with a temporal classifier, and injects them into downstream VideoLLM inference via structured prompting. 
Experiments demonstrate improved motion recognition and more camera-aware model responses, highlighting geometry-driven cue extraction and structured prompting as practical steps toward a camera-aware VideoLLM and VLA system. Dataset and benchmark are available at \footnote{\url{https://huggingface.co/datasets/fengyee/camera-motion-dataset-and-benchmark}}.

\end{abstract}

%% file: sec/1_Introduction.tex
\section{Introduction}
\label{sec:intro}
Video-capable vision-language models (VideoLLMs) have improved substantially on high-level video semantics, including recognition of objects, actions, and narrative events across diverse video lengths~\cite{tang2025video}.
However, an essential component of video meaning, especially in edited content such as films, TV series, and online compilations, lies not only in \emph{what} appears in the frames, but also in \emph{how} it is captured.
Camera motion (\eg, \texttt{\small pan}, \texttt{\small tilt}, and \texttt{\small dolly}) is a core cinematographic device that guides attention, reveals spatial layout, and communicates the author's intent~\cite{lin2025towards}.
Accordingly, camera motion is both central to film grammar and useful for camera-aware description, layout-oriented retrieval, and spatial reasoning.

Despite its importance, current VideoLLMs remain unreliable in recognizing fine-grained camera-motion primitives.
Camera motion is a spatiotemporal geometric signal that is not localized to any single frame, and it is easily confounded by object motion, cuts, and motion blur.
Consequently, models with strong frame-level perception may still fail to model the camera as the source of the visual stream.
Moreover, as we later show, common VideoLLM pipelines compress visual tokens with network depth, which can attenuate motion-sensitive cues.
This gap is unsurprising, as most large-scale video captioning and VQA corpora lack explicit supervision for camera motion.

To study this gap under controlled conditions, we introduce \textbf{CameraMotionDataset}, a synthetic dataset of 12k within-shot segments annotated with fine-grained camera-motion labels. A \emph{shot} is a temporally contiguous sequence of frames captured by a single camera without cuts; we focus on non-overlapping 1-second segments within each shot.
Each segment is annotated with a set of camera-motion labels drawn from a fixed taxonomy of 15 atomic motions. Built on this dataset, \textbf{CameraMotionVQA} provides a multiple-choice benchmark that evaluates open-source VideoLLMs under a standardized VQA protocol.

A central hypothesis of this work is that reliable camera-motion cues can be derived from models with strong geometric and 3D reasoning capabilities \emph{without modifying} the VideoLLM backbone.
Specifically, we use a frozen 3D foundation model (VGGT~\cite{wang2025vggt}) to extract camera cues, train a lightweight temporal classifier to predict constrained motion primitives, and inject the predictions into downstream VideoLLMs via structured prompting.

\begin{figure*}[t]
\centering
\includegraphics[width=\textwidth]{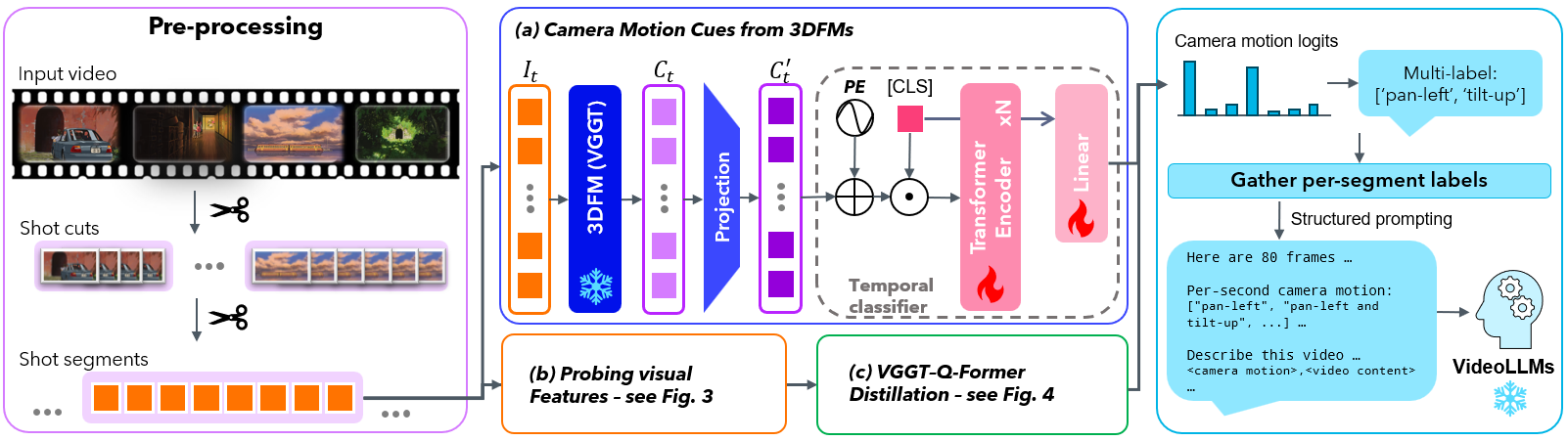}
\vspace{-0.6cm}
\caption{\textbf{Overall pipeline.} Camera cues are extracted from a frozen 3DFM (VGGT) and passed to a Transformer-based temporal classifier to predict camera-motion primitives, and per-second motions are injected as a structured prompt field for VideoLLMs without modifying VideoLLM weights. For clarity, the probing and distillation pipelines are shown separately in \cref{fig:probe_schematic} and \cref{fig:distill_schematic}.}\vspace{-0.5cm}
\label{fig:pipeline}
\end{figure*}

Beyond performance gains, we seek to understand \emph{where} camera-motion information is lost in typical VideoLLM pipelines. By probing intermediate vision-encoder features with Q-Former-style query tokens~\cite{li2023blip}, we observe that camera-motion cues become progressively less recoverable from shallow to deeper blocks, consistent with our hypothesis on token compression and the lack of explicit motion supervision. These findings motivate the use of geometry-derived camera cues from 3DFMs as an external, plug-in signal. Our contributions are threefold:
\begin{itemize}
\item \textbf{CameraMotionDataset and CameraMotionVQA.} We derive a shot-consistent dataset with fine-grained camera-motion labels from an existing synthetic corpus with precise camera parameters, and construct a multiple-choice VQA benchmark for evaluating open-source VideoLLMs.
\item \textbf{Camera-cue assisted motion recognition and structured prompting.} We propose a model-agnostic pipeline that extracts camera cues from a frozen 3DFM (VGGT), predicts motion primitives with a lightweight temporal classifier, and injects them into frozen VideoLLM inference via structured prompting (\cref{fig:pipeline}).
\item \textbf{Probing-based diagnosis.} We analyze where and why camera-motion cues are lost in VideoLLM vision encoders, helping explain failure modes on CameraMotionVQA and motivating our geometry-aware cue injection.
\end{itemize}

%% file: sec/2_Relatedwork.tex
\section{Related Work}
\label{sec:related_work}

\subsection{Camera motion in cinematic video understanding}
\label{sec:rw_benchmarks}

Cinematography depends on both \emph{what} is shown and \emph{how} it is filmed (camera, framing, composition). Recent benchmarks evaluate film-grammar attributes beyond generic video QA. CameraBench~\cite{lin2025towards} defines a cinematographer-informed motion taxonomy, revealing confusions between extrinsic motion (\eg, \texttt{\small dolly}) and intrinsic change (\eg, \texttt{\small zoom}). CineTechBench~\cite{wang2025cinetechbench}, ShotBench~\cite{liu2025shotbench}, and VidComposition~\cite{tang2025vidcomposition} extend evaluation to multi-attribute cinematography, while Shot-by-Shot~\cite{xie2025shot} shows shot-level cues can steer description generation via prompting.

Existing datasets range from per-frame cinematic attributes~\cite{savardi2021cinescale,savardi2023cinescale2,rao2020unified,li2023lightweight,helm2022histshot} to geometry-grounded supervision: SpatialVID~\cite{wang2025spatialvid} provides per-frame depth and pose-derived instructions, and OmniTr~\cite{yang2025omnicam} encodes motion as parameterized programs linking language to trajectories. However, none of them explicitly offer geometry-consistent motion primitives at short-segment granularity.

Camera motion also serves as a control signal in video generation. Recent models~\cite{bai2025recammaster,liu2025posemaster,he2024cameractrl,he2025cameractrl2,bahmani2025ac3d,yu2025trajectorycrafter,luo2025camclonemaster,wang2024motionctrl,yang2024directavideo,shuai2025freeform} condition generation on explicit trajectories, camera embeddings, or 6-DoF poses; MotionCtrl~\cite{wang2024motionctrl} and Direct-a-Video~\cite{yang2024directavideo} decouple camera and object motion control, while others manipulate object poses~\cite{li2025anyi2v,qin2025scenedesigner} or survey multimodal-guided editing~\cite{shuai2024survey}. ReCamMaster~\cite{bai2025recammaster} releases \emph{MultiCamVideo Dataset}, from which we derive our benchmark. These generation-oriented parameterizations also motivate reliable \emph{recognition} of fine-grained primitives under compositional and exclusivity constraints~\cite{he2025cameractrl2}.

At the \emph{object} level, MeViS~\cite{ding2023mevis,ding2025mevis} benchmarks referring segmentation with motion expressions, MOSEv2~\cite{ding2025mosev2} addresses complex-scene video object segmentation (VOS), and MOVE~\cite{ying2025move} adds motion-guided few-shot VOS. These works target object rather than camera motion, but share the motivation to move beyond appearance-based features.

\subsection{Vision-language models and VideoLLMs}
\label{sec:rw_videollm}
VideoLLMs extend image VLMs to video by encoding sampled frames, compressing visual tokens, and conditioning an LLM for instruction following, captioning, and QA. This ``frame aggregation + LLM reasoning" paradigm captures high-level semantics, but token compression can attenuate subtle temporal cues, including camera motion.

Open-source VideoLLMs can be organized by their temporal design and videoLLM interface. Strong-backbone models (\eg, Qwen2.5-VL~\cite{bai2025qwen2} and InternVL~\cite{chen2024internvl}) strengthen the vision tower and time/position encoding to better support long-context video. Connector-based models~\cite{zhang2024llava, xu2024pllava, li2024videochat} keep the LLM mostly fixed and add lightweight temporal modules with aggressive token reduction to fit multi-frame inputs. Instruction-tuned assistants~\cite{cheng2024videollama,zhang2025videollama} emphasize curated video-instruction data and post-training to improve temporal reasoning without major architectural changes. Video-native systems~\cite{wang2024internvideo2, yuan2025tarsier2, hong2024cogvlm2} use stronger video encoders and/or video-centric pretraining to improve spatiotemporal representations.

Bridging modules offer a parameter-efficient way to integrate video into LLMs. Query bottlenecks (\eg, Q-Former) summarize frame features into a small set of query tokens, improving efficiency over dense patch tokens~\cite{li2023blip,zhang2023video}. However, cinematography-centric benchmarks still show weak fine-grained camera-motion recognition, suggesting recent progress is driven more by sampling, grounding, and alignment than explicit camera-motion representation.

\subsection{3D foundation models and geometric cues for VideoLLMs}
\label{sec:rw_3dfm}
3D foundation models (3DFMs) learn transferable 3D representations and infer geometric attributes from visual input. VGGT~\cite{wang2025vggt} is a feed-forward geometry transformer that jointly predicts camera parameters, depth/point maps, and tracking cues from one or multiple views, exposing camera-aware tokens as geometric cues. Follow-up work improves efficiency~\cite{shu2025litevggt}, and learning-based SfM systems amortize reconstruction robustness~\cite{elflein2025light3r,cin2025anymap}, expanding practical sources of camera geometry.

Recent efforts connect geometric priors to vision-language reasoning via explicit geometry branches~\cite{zheng2025learning} or distillation from frozen 3DFMs into compact tokens~\cite{guo2025glad,li2025visual}. VLM-3R~\cite{fan2025vlm3r} takes a tranable approach, augmenting VLMs with instruction-aligned 3D reconstruction that fuses per-view camera tokens with appearance features through end-to-end training. In contrast, our pipeline injects camera cues via structured prompting without modifying VideoLLM weights, making it model-agnostic and complementary to trainable approaches like VLM-3R.

\noindent\textbf{Summary.} Benchmarks position camera motion at the intersection of geometry and semantics: although datasets provide explicit trajectories, VideoLLMs struggle to recover fine-grained motion primitives. We therefore leverage off-the-shelf 3DFMs to extract camera cues, predict constraint-consistent motion labels, and inject them into frozen VideoLLMs via structured prompting. %

%% file: sec/3_Method.tex
\section{Method}\label{sec:method}

\subsection{Problem overview}
\label{sec:method:overview}
Given an input video, we split it into shots via an off-the-shelf detector (\eg, Shot-by-Shot~\cite{xie2025shot}) and divide each shot into non-overlapping 1-second segments. Each segment is assigned a multi-hot label $\mathbf{y}\in\{0,1\}^{K}$ over $K{=}15$ atomic primitives; co-occurrence is allowed (\eg, \texttt{\small pan-left} + \texttt{\small tilt-up}) but mutually exclusive pairs (\eg, \texttt{\small pan-left} vs.\ \texttt{\small pan-right}) are forbidden.
Our module is lightweight and plug-and-play: a frozen 3DFM extracts camera cues, a temporal classifier predicts motion primitives, and the per-second predictions are injected into a structured prompt for downstream VideoLLMs (see \cref{fig:pipeline}).
We adopt VGGT as the 3DFM because it produces per-frame \emph{camera tokens} in a single forward pass, encoding pose and dynamics in a unified coordinate system---cues that we find are not well-preserved by VideoLLM vision encoders.

\begin{figure*}[t]
\centering
\includegraphics[width=0.85\linewidth]{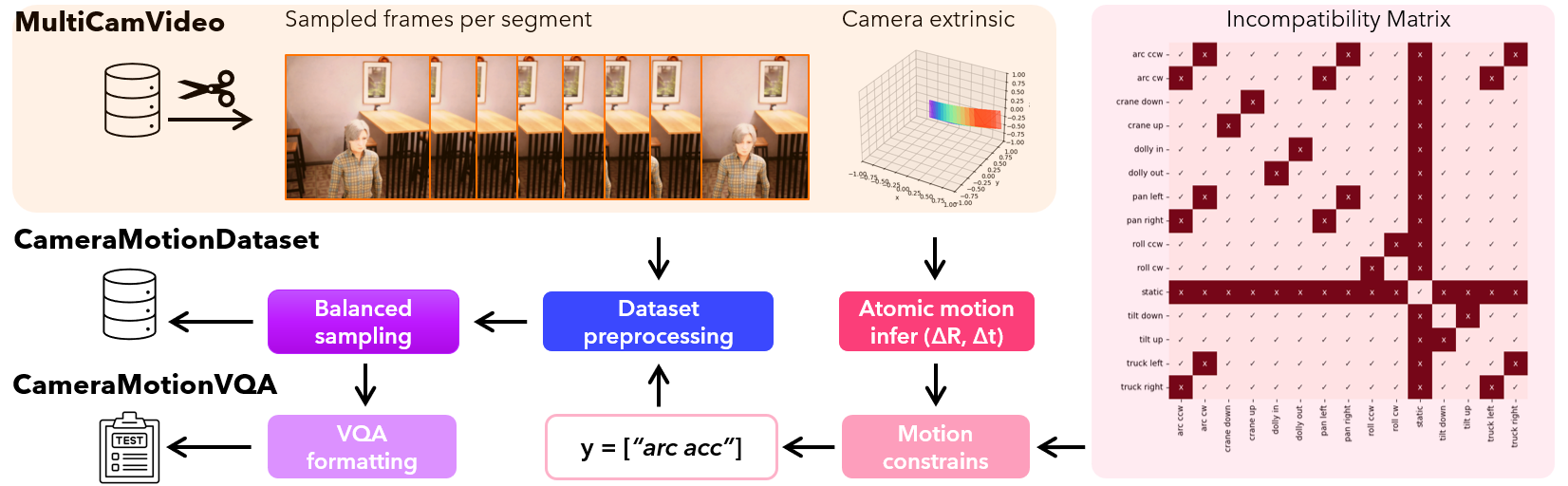}
\vspace{-0.2cm}
\caption{\textbf{Flow chart for dataset/benchmark construction.} From MultiCamVideo~\cite{bai2025recammaster}, video clips and camera extrinsics are preprocessed (split, resized, and normalized) and labeled with several motion constraints (\eg, incompatibility). A subset of shot segments is sampled based on motion primitives to balance classes, before storing as CameraMotionDataset and formulated into CameraMotionVQA records.}
\label{fig:benchmark}\vspace{-0.4cm}
\end{figure*}

\subsection{CameraMotionDataset construction and CameraMotionVQA benchmark}
\label{sec:method:dataset}

We build a labeled camera-motion dataset from the \emph{MultiCamVideo} dataset introduced in ReCamMaster~\cite{bai2025recammaster}.
MultiCamVideo contains photorealistic dynamic scenes rendered in Unreal Engine 5~\cite{UE5}, with dense frame-wise camera calibration (intrinsics) and camera poses (extrinsics). It covers 13.6K dynamic scenes and includes 136K videos with 112K distinct camera trajectories; videos are rendered at 15 fps, and trajectories are explicit camera extrinsic sequences.

As shown in \cref{fig:benchmark}, each video is divided into non-overlapping 1-second segments to ensure camera motion labels are accurately aligned with actual camera changes. For each segment, we uniformly sample $T$=8 frames and resize to $336\times336$, which provides a favorable accuracy--efficiency trade-off for camera motion recognition.

\noindent\textbf{Camera motion label from extrinsic params.} We assign camera motion labels by analyzing frame-wise camera extrinsic matrices, following the taxonomy and operational definitions in CameraBench~\cite{lin2025towards}.
Concretely, we compute per-segment translation and rotation changes (\eg, yaw/pitch/roll deltas and forward/backward translation) and map them to motion primitives (\eg, \texttt{\small pan-left}, \texttt{\small tilt-down}, \texttt{\small dolly-in}) using thresholded pattern matching in pose dynamics. 
Compound motions are produced when multiple primitives are detected simultaneously (\eg, \texttt{\small arc-clockwise}, \texttt{\small dolly-in}).
With precisely controlled camera motion, our labeling approach provides sufficiently accurate annotation results for each segment. To validate label quality, we conduct a human verification study on 720 randomly selected segments and observe 93\% agreement.

We further observe substantial class imbalance.
Thus, we apply stratified sampling to construct a balanced subset while preserving diversity across atomic and compound-motion classes, yielding 12,274 segments in total. For \emph{CameraMotionDataset}, a standard train/val/test split is created by 80\%/10\%/10\% folds. Camera motion taxonomy and labeling procedure, and dataset re-balancing details are provided in the supplementary.

\noindent\textbf{Atomic primitives, constraints, and canonicalization.}
We define 15 atomic primitives and represent each segment label as a multi-hot vector.
To enforce physically and semantically valid combinations, we define a symmetric incompatibility matrix $\mathbf{M}\in\{0,1\}^{K\times K}$, where $\mathbf{M}_{ij}=1$ indicates that primitives $i$ and $j$ cannot co-occur (see \cref{fig:benchmark}).
Label-set canonicalization limits the maximum number of primitives, uses deterministic ordering and de-duplication, for consistent evaluation and prompt serialization. 

\noindent\textbf{CameraMotionVQA benchmark.} To benchmark off-the-shelf VideoLLMs under a unified protocol, we construct a multiple-choice VQA benchmark, named CameraMotionVQA.
Each question corresponds to a 1-second clip and provides four candidate answers: one ground-truth motion label set and three \emph{distractors}.
Distractors are sampled to have similar label complexity as the ground-truth (number of primitives), avoiding trivial answer-length bias. In addition, all candidates are valid under the incompatibility constraints. VQA templates are included in the supplementary.

\cref{tab:positioning} positions our dataset and benchmark against existing cinematic benchmarks and video datasets. Based on this comparison, we argue that it is necessary to create a camera motion-focused dataset with trustworthy labels derived deterministically from extrinsic params and sufficient scale to train a robust modern classifier.

\begin{table*}[t]
\centering
\footnotesize
\setlength{\tabcolsep}{4pt}
\renewcommand{\arraystretch}{1.15}
\caption{\textbf{Positioning of our datasets against prior benchmarks and datasets.} ``Motion-level'' indicates the granularity of camera supervision (primitive vs. composition of motions). Our CameraMotionDataset provides constrained primitive-level supervision with explicit synthetic camera parameters, while CameraMotionVQA converts the same 1s within-shot segments into a multiple-choice evaluation protocol. A full comparison with detailed annotation structure and intended usage is provided in the supplementary material.}\label{tab:positioning}\vspace{-0.2cm}
\begin{tabular}{l|c|c|c|c|c|c}
\hline
 & \textbf{Motion-level} & \textbf{Granularity} & \textbf{Temporal unit} & \textbf{QA protocol} & \textbf{Intended use} & \textbf{Camera params} \\
\hline
CameraBench~\cite{lin2025towards} & Primitive & Multi-label & Short clips & Caption, Yes/No QA & Benchmark + fine-tune & No \\
CineTechBench~\cite{wang2025cinetechbench} & Composition & Coarse & Frames / short clips & Caption, MCQ & Benchmark & No \\
VidComposition~\cite{tang2025vidcomposition} & Composition & Coarse & Compiled videos & MCQ & Benchmark & No \\
CineScale2~\cite{savardi2023cinescale2} & - & - & Frames & No & Training dataset & No \\
\hline
\textbf{CameraMotionDataset} & \textbf{Primitive} & \textbf{Multi-label} & \textbf{1s within-shot} & No & \textbf{Training dataset} & \textbf{Yes} \\
\textbf{CameraMotionVQA} & \textbf{Primitive} & \textbf{Multi-label} & \textbf{1s within-shot} & \textbf{MCQ} & \textbf{Benchmark} & No \\
\hline
\end{tabular}\vspace{-0.2cm}
\end{table*}

\subsection{Camera motion cues from 3DFMs}
\label{sec:method:vggt}
Camera motion is fundamentally geometric, arising from coherent changes in camera pose over time.
Although modern VideoLLMs are trained for semantic alignment and temporal coherence, geometric camera-motion cues are not reliably preserved in their intermediate vision representations (Sec.~\ref{sec:method:probing}).
To compensate for this missing geometric signal, we use a 3DFM as an external camera motion cue extractor.

Our framework is agnostic to the specific 3DFM: any model that can provide per-frame camera descriptors (\eg, pose-related tokens or camera parameter estimates) can be plugged into our pipeline. VGGT~\cite{wang2025vggt} produces camera tokens in a single forward pass and has been shown to encode camera pose and motion dynamics.
Given a 1-second segment with $T$ sampled frames, we run VGGT on each frame and obtain a camera token sequence $\{\mathbf{c}_{t}\}_{t=1}^{T}$, where $\mathbf{c}_{t}\in\mathbb{R}^{2048}$.
These camera tokens serve as the input to our temporal classifier, as shown in \cref{fig:pipeline}.

\subsection{Constraint-regularized motion classifier}
\label{sec:method:classifier}

We train a lightweight classifier to map camera tokens $\mathbf{c}_t$ to constrained multi-label camera-motion predictions.
A linear projection $\mathbf{W}_{p}$ is applied to form an information bottleneck: VGGT tokens may encode rich camera-related factors, and the projection stabilizes training by distilling the subset most relevant to motion prediction.
Then, add a sinusoidal positional encoding and prepend a learnable \texttt{\small [CLS]} token.
The resulting sequence is processed by an $L$-layer Transformer encoder, and logits are predicted from the final \texttt{\small [CLS]} token (\ie, $\mathbf{Z}_{L}[0,:]$) by a linear projection $\mathbf{W}_{o}$:
\begin{align}
\mathbf{z}_{t} &= \mathrm{PE}(\mathbf{W}_{p}\mathbf{c}_{t}), \quad t=1,\ldots,T,\\
\mathbf{Z}_{0} &= \left[\mathbf{z}_{\mathrm{cls}},\mathbf{z}_{1},\ldots,\mathbf{z}_{T}\right],\\
\mathbf{Z}_{L} &= \mathrm{TransformerEnc}(\mathbf{Z}_{0}),\\
\mathbf{s} &= \mathbf{W}_{o}\mathbf{Z}_{L}[0,:] + \mathbf{b}_{o},
\end{align}
where $\mathbf{s}\in\mathbb{R}^{K}$ are the output logits and $p_k = \mathrm{sigmoid}(s_k)$ denote per-class probability. Following the constrained multi-label formulation in Sec.~\ref{sec:method:overview}, we optimize a binary cross-entropy loss with two regularizations:
\begin{align}
\mathcal{L}_{\mathrm{bce}} &= -\sum_{k=1}^{K}\Big( y_{k}\log p_{k} + (1-y_{k})\log(1-p_{k}) \Big), \\
\mathcal{L}_{\mathrm{inc}} &= \sum_{1\le i < j \le K} \mathbf{M}_{ij} p_{i} p_{j}, \\
\mathcal{L}_{\mathrm{card}} &= \max\Big(0,1-\sum_{k=1}^{K} p_{k}\Big)^{2}+\max\Big(0,\sum_{k=1}^{K} p_{k}-3\Big)^{2}
\end{align}
where $\lambda_{\mathrm{inc}}$ and $\lambda_{\mathrm{card}}$ control the strengths of incompatibility and cardinality regularization. In our implementation, we use $\lambda_{\mathrm{inc}}=1.0$ and $\lambda_{\mathrm{card}}=1.0$ by default.

At inference, we threshold probabilities at $\tau{=}0.5$ and enforce constraints by removing mutually exclusive primitives within each primitive group and applying canonicalization.

\subsection{Motion injection via structured prompting}
\label{sec:method:prompt}
As a training and fine-tuning free approach, camera motion cues are injected via structured prompting, leaving VideoLLM weights unchanged. For a shot consisting of $S$ one-second segments, each motion label is predicted and serialized as a short string (\eg, \texttt{\small static} or \texttt{\small pan-left and tilt-up}), and these strings are concatenated into a per-shot list:
\texttt{\small Per-second camera motion: $[m_1, m_2, \ldots, m_S]$}.
We prepend this motion list to the user instruction to provide an explicit temporal scaffold.
Empirically, this conditioning might improve the temporal grounding of generated descriptions and reduce camera-motion hallucinations, as the model can align content changes with the provided motion sequence (\cref{fig:qual_prompt}). Our final prompt template is:
\begin{quote}
\footnotesize\ttfamily
Here are [$N$] consecutive video frames. They are evenly sampled at a frame rate of [$r$] FPS.\\
Per-second camera motion: $[m_1, m_2, \ldots]$.\\
Describe this video using the filmmaker's language, highlighting lighting, framing, composition, and especially camera usage that connects different frames.\\
For example: "At the beginning, <video content>; then <camera motion>, <video content>; ...; finally, <camera motion>, <video content>."
\end{quote}

\subsection{Probing motion sensitivity via Q-Former}
\label{sec:method:probing}

\begin{figure}[t]
\centering
\includegraphics[width=\linewidth]{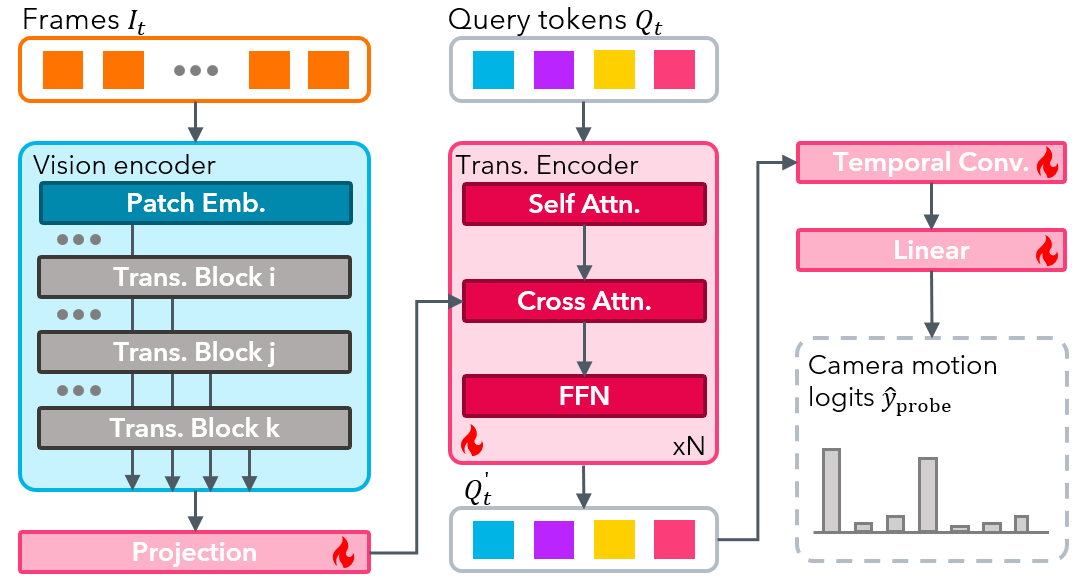}
\vspace{-0.6cm}
\caption{\textbf{Probing experiment schematic.} Query tokens $Q_t$ gather camera motion-related information from the projected intermediate visual features of the frozen vision encoder. Camera motion logits are predicted from the temporal convolution output of the transferred vision tokens $Q_t'$.}
\label{fig:probe_schematic}\vspace{-0.5cm}
\end{figure}

Q-Former~\cite{li2023blip} style probing provides an effective ``readout'' mechanism: a small set of learnable query tokens can extract task-relevant information from high-dimensional frozen visual tokens via cross-attention, enabling a parameter-efficient diagnostic of what information is present in the representation. As shown in \cref{fig:probe_schematic}, we probe on Qwen2.5-VL~\cite{bai2025qwen2}, whose vision encoder is a dynamic-resolution ViT with window attention blocks and periodic full-attention blocks at indices $\{7,15,23,31\}$. Thus, we extract visual tokens from patch embedding and the full-attention blocks.
This pipeline consists of: (i) a linear projection that bottlenecks the frozen visual tokens to the query dimension; (ii) 4 learnable query tokens processed by 2 Transformer Encoders; (iii) a 1D temporal convolution over the query tokens to produce a single classification token; and (iv) a linear classifier to predict multi-label logits.
The probe is trained with the same loss in Sec.~\ref{sec:method:classifier}.

\begin{figure}[t]
\centering
\includegraphics[width=\linewidth]{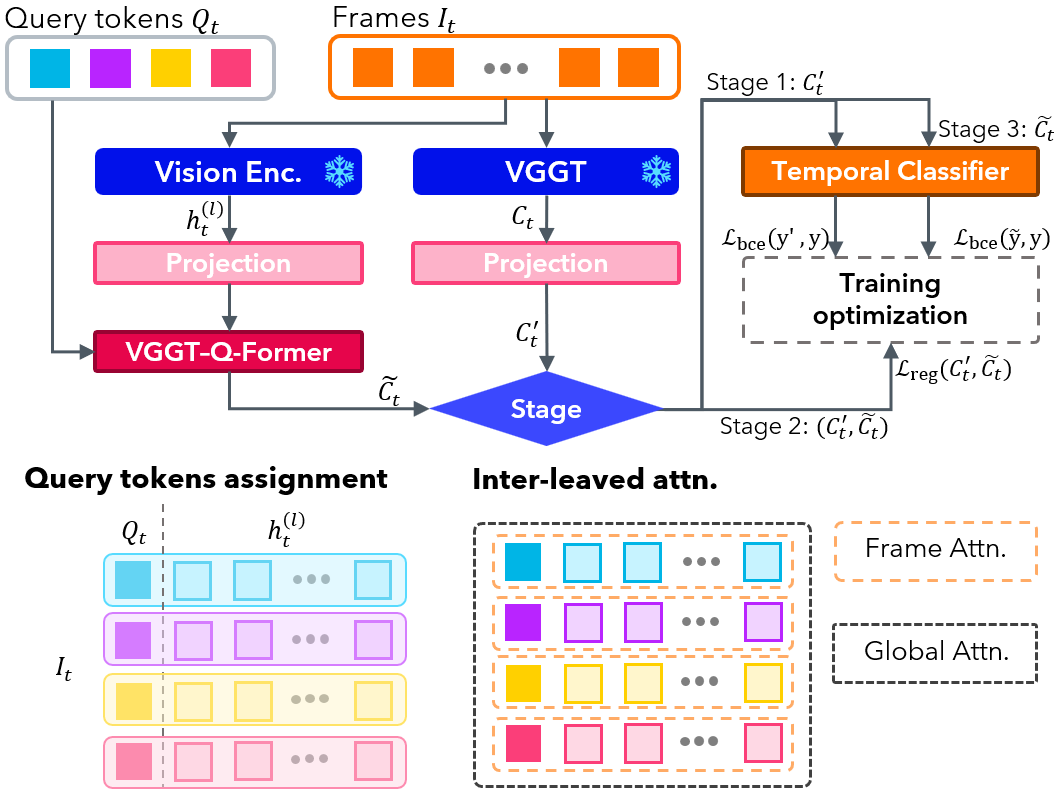}
\vspace{-0.6cm}
\caption{\textbf{VGGT--Q-Former schematic.} Camera tokens and visual tokens are both bottlenecked by a projection layer. Query tokens gather camera motion-related information via interleaved local-/global-attention blocks and regress the projected camera tokens to distill the camera perception capability of VGGT. Branch annotated with Stage $i$ indicating different training optimization objectives and flows of tokens.}
\label{fig:distill_schematic}\vspace{-0.5cm}
\end{figure}

\subsection{VGGT--Q-Former distillation}
\label{sec:method:distill}
VGGT provides high-quality camera tokens but is computationally heavy (1.2B params).
As shown in \cref{fig:distill_schematic}, to reduce inference cost, we propose to distill VGGT's camera perception by a lightweight Q-Former-based student. Specially, the student follows the same \emph{interleaved local- and global-attention} as VGGT for camera reasoning, enabling temporally grounded yet context-aware representations.

\noindent\textbf{Q-Former with interleaved local-frame and global attention.}
We assign one learnable query token to each (temporally indexed) feature map and have processing blocks that alternate between local (frame) attention and global attention.
In local attention, each query attends only to the visual tokens of its assigned frame, encouraging temporally grounded camera cues. In global attention, all queries attend jointly across frames, enabling cross-frame aggregation of camera dynamics.
As in VGGT, query tokens $Q_t$ have 256 dimensions, and the outputs from the final local and global blocks are concatenated to form the final distillation result $\tilde{c}_t$ of dim 512.

\noindent\textbf{Three-stage progressive training.}
As illustrated in \cref{fig:distill_schematic}, a progressive training strategy is adopted: (1) train the motion classifier on projected VGGT tokens; (2) train the distiller (VGGT--Q-Former) by regressing the projected VGGT tokens using mean squared error; and (3) jointly fine-tune the VGGT--Q-Former and classifier.
The regression loss is
\begin{equation}
\mathcal{L}_{\mathrm{reg}} = \sum_{t=1}^{T} \big|\widetilde{\mathbf{c}}_{t} - \mathbf{c}^{\prime}_{t}\big|_{2}^{2},
\end{equation}
where $\mathbf{c}^{\prime}_{t}$ denotes the projected VGGT token and $\widetilde{\mathbf{c}}_{t}$ denotes the distilled token.

At inference, VGGT is replaced by VGGT--Q-Former, and the same temporal motion classifier is used to get labels for each segment. In Sec.~\ref{sec:exp_distill}, we report the trade-off between efficiency and accuracy of this distillation approach.

%% file: sec/4_Experiment.tex
\section{Experiments}
\label{sec:experiments}

\subsection{Experimental setup}
\label{sec:exp_setup}

All experiments are conducted on a single NVIDIA RTX A6000 GPU. We evaluate camera-motion recognition on \textbf{CameraMotionDataset} and \textbf{CameraMotionVQA} as detailed in Sec.~\ref{sec:method:dataset}.
CameraMotionVQA formats each segment as a 4-way multiple-choice question, and models are evaluated using answer accuracy.
For evaluating the camera motion multi-label task, we report \textbf{instance accuracy} (exact match of all labels), \textbf{Macro-F1} across motion primitives, and \textbf{Weighted-F1} weighted by label frequency.

\subsection{How well do off-the-shelf VideoLLMs recognize camera motion?}

\label{sec:exp_vlm_baselines}
\begin{figure}
\centering
\includegraphics[width=\linewidth]{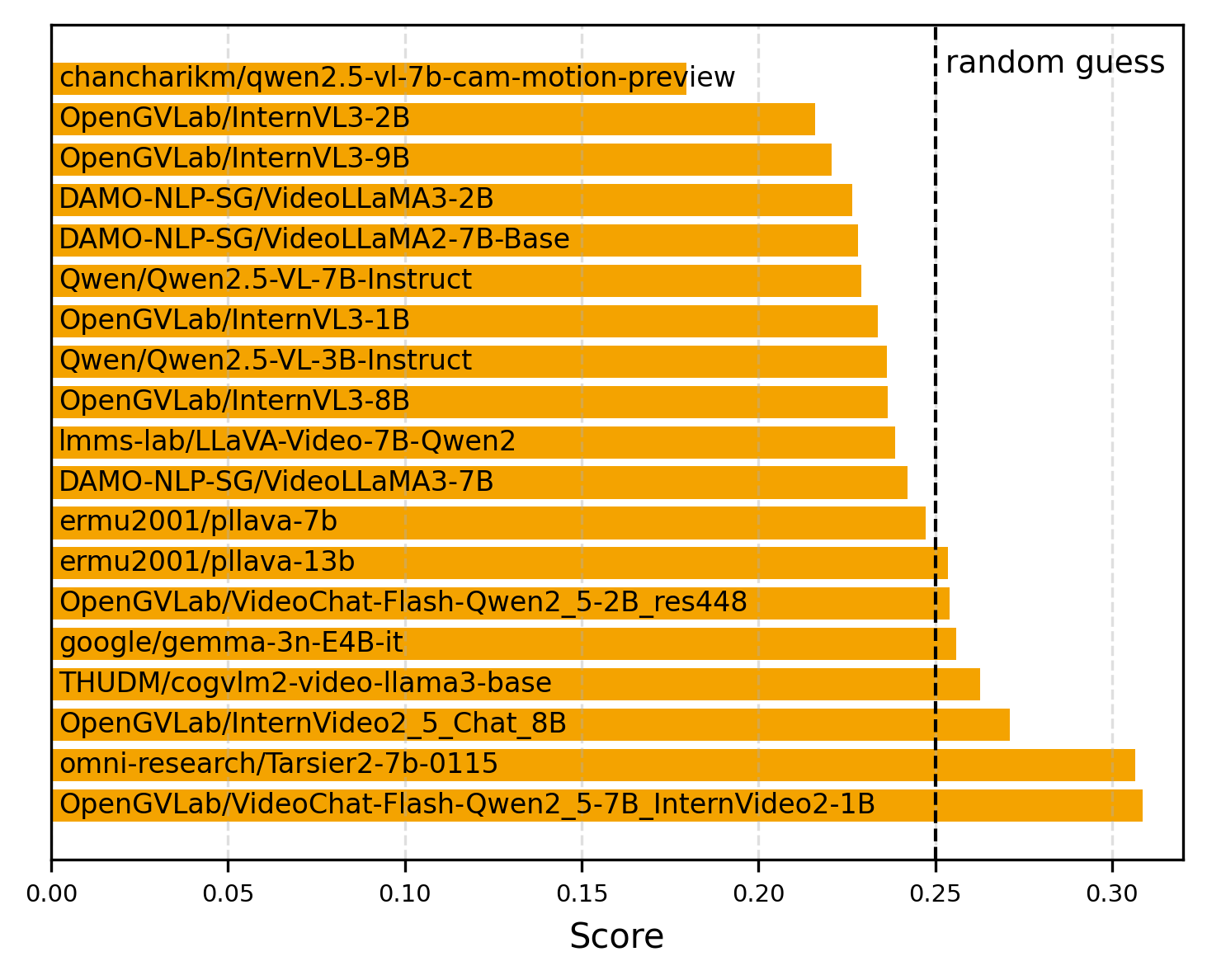}
\vspace{-0.8cm}
\caption{\textbf{Off-the-shelf VideoLLM performance on CameraMotionVQA.} Horizontal bars report overall multiple-choice accuracy. All models, labeled by their {Hugging Face model names}, use an identical frame input and VQA prompt template.}
\label{fig:vlm_baselines}\vspace{-0.5cm}
\end{figure}

As described in Sec.~\ref{sec:method:dataset}, we evaluate diverse off-the-shelf VideoLLMs under the unified CameraMotionVQA protocol. %
\cref{fig:vlm_baselines} shows that most models perform near the random-guess rate (25\%), revealing a substantial \emph{camera-motion blindness} gap. Notably, a structured fine-tuning baseline from CameraBench~\cite{lin2025towards} performs worse than its off-the-shelf counterpart (Qwen2.5-VL). Qualitatively, models frequently confuse geometrically similar primitives (e.g., \texttt{\small truck} vs. \texttt{\small pan}) and produce inconsistent directional predictions (e.g., \texttt{\small pan-left} vs. \texttt{\small pan-right}) in the presence of salient object motion.

We attribute this gap to the lack of explicit camera-motion supervision in VideoLLM training data, where representations are optimized for semantic alignment and temporal reasoning rather than precise 3D geometric change. Our probing results (Sec.~\ref{sec:exp_probing}) support this claim and motivate injecting explicit geometric camera cues.

\subsection{Camera motion recognition from 3D foundation-model cues}
\label{sec:exp_vggt_classifier}

We run a frozen 3D foundation model (VGGT) on $\{I_t\}_{t=1}^{8}$ and extract per-frame camera cues (camera tokens $C_t\in\mathbb{R}^{2048}$). A lightweight temporal classifier (Sec.~\ref{sec:method:classifier}) consumes the cue sequence and predicts constrained multi-label motions. Unless otherwise stated, VGGT is frozen, and we train a shallow Transformer classifier with 4 encoder blocks and 8 attention heads. Each camera token is projected to $C_t'\in\mathbb{R}^{512}$; a learnable \texttt{\small [CLS]} token is prepended; the final \texttt{\small [CLS]} embedding is mapped to $K$ logits. In the supplementary, we ablate the number of encoder blocks, attention heads, and hidden size, and find that the current setting achieves a favorable accuracy--compute trade-off.

\begin{table}[t]
\centering
\small
\setlength{\tabcolsep}{2pt}
\caption{\textbf{Multi-label camera-motion recognition results on the test split of CameraMotionDataset.} \textit{Inst. Acc.}: exact multi-label matching accuracy; \textit{Macro-F1}: class-averaged F1 score; \textit{Weighted-F1}: sample frequency-weighted F1 score.} 
\label{tab:main_results}\vspace{-0.2cm}
\begin{tabular}{lccc}
\toprule
\textbf{Method} & \textbf{\footnotesize Inst. Acc.}$\uparrow$ & \textbf{\footnotesize Macro-F1}$\uparrow$ & \textbf{\footnotesize Weighted-F1}$\uparrow$ \\
\midrule
VGGT w. constraints & \textbf{0.738} & \textbf{0.87} & \textbf{0.92} \\
VGGT w/o. constraints & 0.572 & 0.79 & 0.84 \\
VGGT--Q-Former & 0.638 & 0.83 & 0.87 \\
Q-Former probing & 0.450 & 0.69 & 0.74 \\
\bottomrule
\end{tabular}\vspace{-0.4cm}
\end{table}

\cref{tab:main_results} shows that VGGT-derived cues with a lightweight classifier substantially outperform off-the-shelf VideoLLM baselines (\cref{fig:vlm_baselines}).
Removing constraint enforcement (\textit{VGGT w/o. constraints}) reduces instance-level accuracy, indicating that modeling axis-wise mutual exclusion improves performance even with strong cues. Per-label prediction results (in the supplementary) show that errors concentrate on rare or ambiguous primitives, consistent with the long-tail distribution.
We also observe unreliable predictions for the \texttt{\small static} class, which is likely out-of-distribution for VGGT, whose reconstruction prior assumes camera motion.
Static segments may require dedicated handling beyond 3DFM priors.

Despite its effectiveness, VGGT cue extraction is costly: the 1.2B-parameter model performs multi-view 3D reasoning over all frames, dominating latency and memory.

\subsection{Probing motion sensitivity in a vision encoder}

\label{sec:exp_probing}
As described in Sec.~\ref{sec:method:probing}, we freeze the Qwen2.5-VL vision encoder and train a small Q-Former-style probe on intermediate features to predict motion labels. This pathway is used \textit{solely} to diagnose representation bottlenecks. The probe comprises 2 Transformer blocks with 8 heads and 4 learnable query tokens. Query tokens and projected features are 768-dimensional.

\begin{figure}[h!]
\centering
\includegraphics[width=0.9\linewidth]{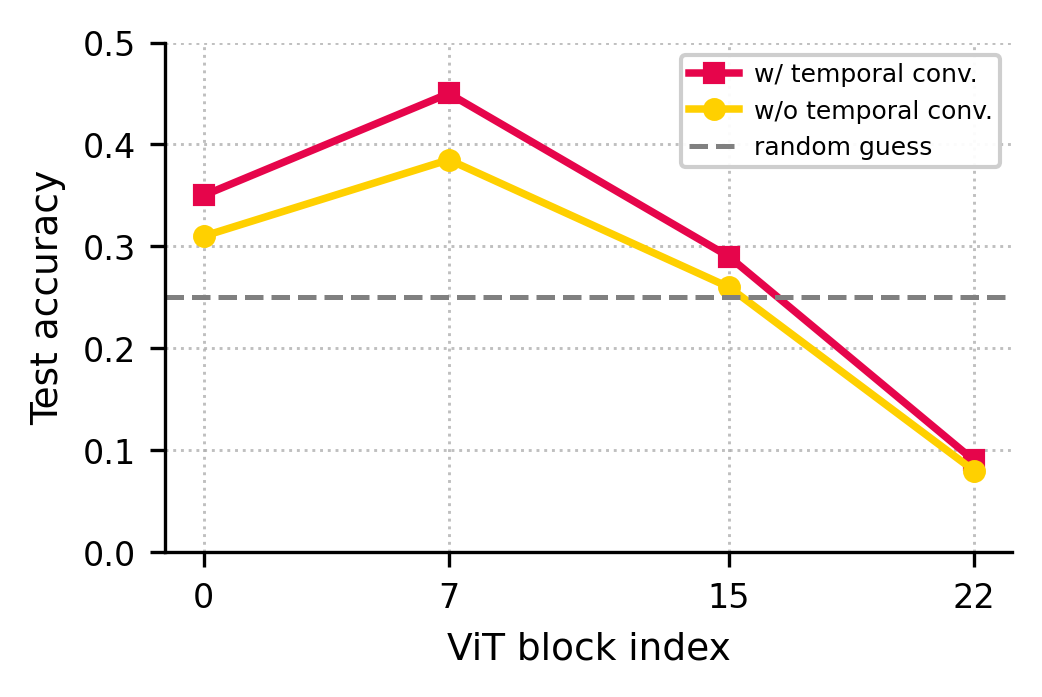}
\vspace{-0.4cm}
\caption{\textbf{Results of probing experiment.} Intermediate features from the frozen vision tower at different Transformer block indices are probed by query tokens. Performance peaks at a shallow intermediate block and degrades in later layers, suggesting that camera-motion cues are not reliably preserved.} \label{fig:probing_layers}\vspace{-0.2cm}
\end{figure}

\begin{figure*}[t]
\centering
\includegraphics[width=1\linewidth]{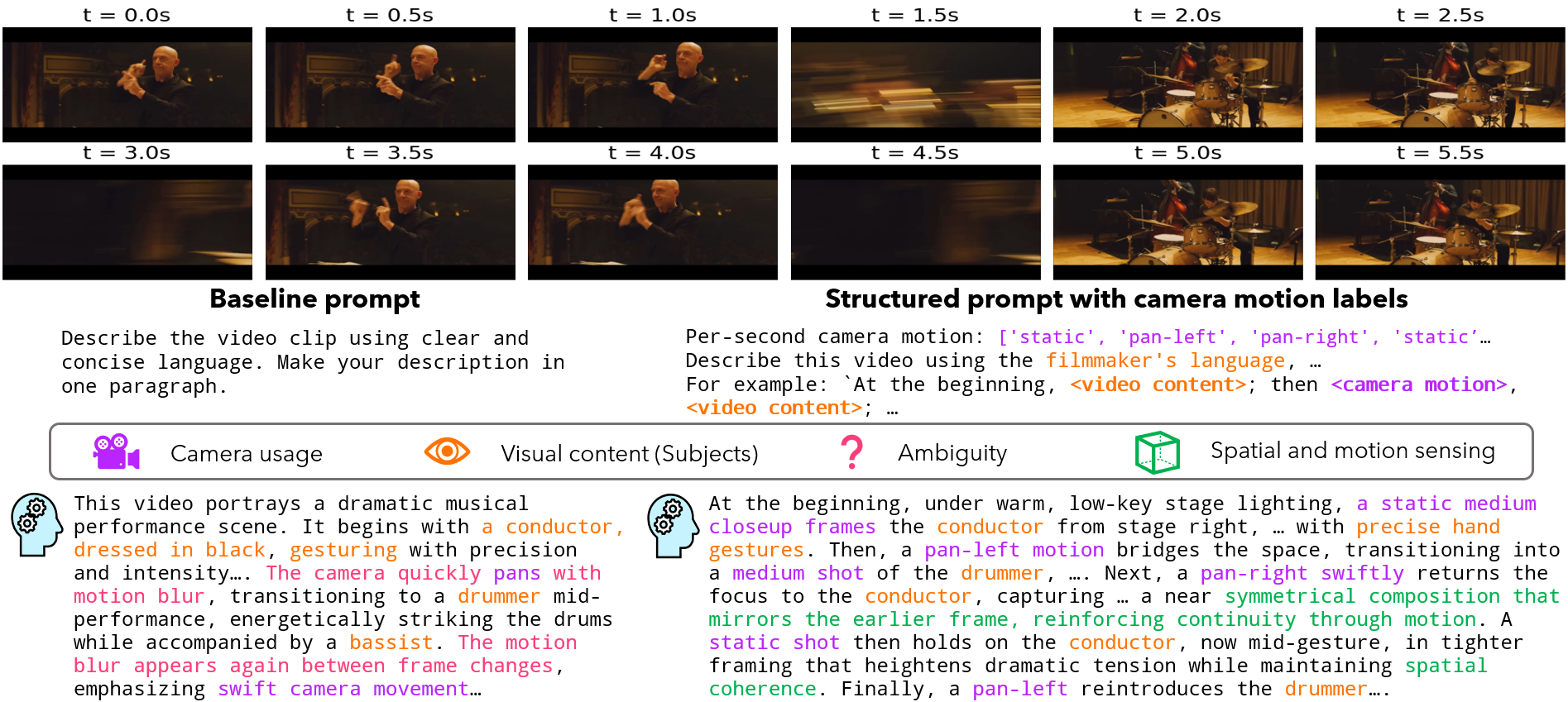}
\vspace{-0.4cm}
\caption{
\textbf{Structured motion header enables camera-aware and temporally grounded descriptions.}
Top: a 5-second clip sampled at 2 fps. Bottom: outputs from the same VideoLLM with different prompts.
The baseline description correctly identifies subjects but produces {\color{VioletRed} ambiguous motion statements}, without specifying {\color{Green} direction or temporal structure}.
With structured motion cues, the model generates explicit \textcolor{mypurple}{camera usage}, maintains {\color{orange} subject} focus, and introduces {\color{Green} spatial-temporal reasoning such as continuity and coherence}.}
\label{fig:qual_prompt}\vspace{-0.2cm}
\end{figure*}

As shown in \cref{fig:probing_layers}, performance peaks at the first full-attention block and declines in later layers. This suggests that camera-motion information is weakly encoded and becomes less accessible as features are optimized for semantic alignment with the language model.
Replacing temporal convolution with average pooling further reduces accuracy, indicating that explicit temporal modeling is necessary to derive camera motion from token sequences.

\subsection{Distilling 3D Camera Cues for efficiency}
\label{sec:exp_distill}

VGGT-based cue extraction is computationally expensive at inference: the 1.2B-parameter backbone performs multi-view 3D reasoning over all frames. To reduce cost, we distill teacher camera tokens into a compact student Q-Former (VGGT--Q-Former) that reuses frozen VideoLLM vision features (Sec.~\ref{sec:method:distill}). We use hidden features $h_t^{l}$ from the 7th block of Qwen2.5-VL, where probing shows higher camera-motion recoverability (\cref{fig:probing_layers}). The student attends to visual tokens and predicts embeddings that regress projected teacher tokens.
We use 4 learnable queries (one per frame group) and follow VGGT's interleaved local/global attention (\cref{fig:distill_schematic}): 2 local blocks (query--frame) and 2 global blocks (cross-frame).
Training proceeds in three stages: (1) train the motion classifier for 50 epochs; (2) train query tokens for 100 epochs to regress teacher tokens; (3) jointly fine-tune the student and classifier for 30 epochs.
We use Adam with a learning rate of 1e-4.

\begin{table}[t]
\centering
\small
\setlength{\tabcolsep}{6pt}
\caption{\textbf{Efficiency comparison of camera-motion recognition pipelines.}
We compare trainable parameter count, peak GPU memory, and inference time throughput for: the full VGGT-based classifier; the distilled VGGT--Q-Former pipeline; and the Q-Former probing experiment. All measurements are conducted on an RTX A6000 with batch size 16 and input resolution $336\times336$.}
\label{tab:efficiency}\vspace{-0.2cm}
\begin{tabular}{lccc}
\toprule
\textbf{Pipeline} &
\multicolumn{1}{c}{\textbf{Params}} &
\multicolumn{1}{c}{\textbf{Peak mem.}} &
\multicolumn{1}{c}{\textbf{Throughput}} \\
&
\multicolumn{1}{c}{\textbf{(M)}} &
\multicolumn{1}{c}{\textbf{(MBs)}} &
\multicolumn{1}{c}{\textbf{(samples/s)}} \\
\midrule
VGGT classifier & 9.47 & 23649.11 & 4.39 \\
VGGT--Q-Former & 9.15 & 9202.63 & 23.36 \\
Q-Former probing & 15.18 & 9232.23 & 25.12 \\
\bottomrule
\end{tabular}\vspace{-0.4cm}
\end{table}

\cref{tab:efficiency} reports inference overhead. Although instance accuracy drops by 8.13\%, distillation substantially reduces end-to-end cost. The distilled model (8.72M \textit{vs.} 1.2B) achieves 5.3$\times$ throughput at 39\% peak memory, offering a favorable accuracy--latency trade-off.
Overall, teacher cues maximize accuracy but are costly, whereas distilled cues improve efficiency at some loss in accuracy.

\subsection{Qualitative results for structured prompting}
\label{sec:exp_prompting}
Our goal is to enable VideoLLMs to produce camera-aware, filmmaker-style descriptions that capture \emph{what} happens and \emph{how} the shot is filmed and evolves over time.
Following Sec.~\ref{sec:method:prompt}, we prepend a per-second motion header to the prompt without modifying model weights.
\cref{fig:qual_prompt} shows a representative example. Under the baseline prompt, the model recognizes the conductor and drummer, but describes motion vaguely (\eg, ``camera quickly pans with motion blur") without direction or temporal structure.
The description mixes possible cuts and camera movement, leading to ambiguity in how frames are connected.

In contrast, with the provided explicit motion cues, the same VideoLLM produces a temporally structured narrative.
It explicitly grounds motion direction (\texttt{\small pan-left} and \texttt{\small pan-right}), describes framing (\texttt{\small static medium close-up}), and adds spatial reasoning (\eg, ``mirrors the earlier frame'', ``spatial coherence''). This suggests motion headers improve motion correctness and encourage geometry-aware, temporally consistent reasoning.

Beyond motion correctness, the header biases the model toward spatial and motion reasoning. The structured cues promote continuity, geometric consistency, and frame-to-frame transitions, reducing generic descriptions.
This suggests that camera motion labels act as geometric priors that steer large VideoLLMs toward more temporally grounded and cinematographically aware reasoning.
Full outputs and other examples are provided in the supplementary.

%% file: sec/5_Conclusion.tex
\section{Conclusion and Discussion}
\label{sec:conclusion}
Camera motion is a fundamental geometric signal that shapes video perception, yet it is rarely modeled explicitly in current VideoLLMs.
In this work, this gap is addressed through a loop of \emph{benchmarking}, \emph{diagnosis}, and \emph{injection}: we introduce a shot-consistent 1-second dataset and a VQA benchmark that formulates fine-grained motion primitives as a constrained multi-label recognition task; we show via probing that camera motion cues are only weakly recoverable from frozen VideoLLM vision features; and we propose a lightweight pipeline that extracts cues from a frozen 3D foundation model (VGGT).
The pipeline predicts constrained motion primitives and injects them via structured prompting \emph{without updating VLM weights}.

Camera cues support applications where \emph{how} a scene is filmed matters in addition to \emph{what}, including descriptive video services (DVS)~\cite{xie2025shot}, media recommendation and retrieval by filmmaking metadata~\cite{savardi2023cinescale2}, camera-aware video attribute or authorship analysis~\cite{li2023lightweight}, and plagiarism detection.
More broadly, our results suggest camera awareness remains a structured supervision problem and there is representation gap in current VideoLLMs. In this work, we propose a plug-and-play solution, in which synthetic camera-controlled data provides scalable supervision, 3DFMs inject geometric priors, and distillation offers a favorable cost--performance trade-off.
Limitations include the synthetic-to-real gap, focus on camera \emph{extrinsic} rather than \emph{intrinsic} changes (\eg, zoom), and only a single 3DFM backbone is explored.
A detailed discussion is provided in the supplementary, including a data augmentation strategy, and a LLM-as-judge evaluation protocol.
Future work will extend to broader camera configurations and 3DFMs, and systematically study how camera cues transfer to downstream video understanding and generation tasks.